%% file: main.tex
\documentclass{vgtc}

\usepackage{enumitem}

\input{preamble}

\vgtcinsertpkg
\usepackage{orcidlink}

\title{LICOA: Learning Instruction-Conditioned Optical Actions for Mobile Spatial Augmented Reality}

\author{Yuchen Deng\orcidlink{0009-0006-3893-955X}\thanks{e-mail: swudyc714@email.swu.edu.cn}\\ %
     \scriptsize Southwest University %
\and Shunkai Zhou\orcidlink{0009-0002-4597-0695}\thanks{e-mail: skzhou25@stu.pku.edu.cn}\\ %
     \scriptsize Peking University %
\and Ruixuan Zhu\orcidlink{0009-0004-9980-0792}\thanks{e-mail: euphorbiamilii@email.swu.edu.cn}\\ %
     \scriptsize Southwest University %
\and Haibin Ling\orcidlink{0000-0003-4094-8413}\thanks{e-mail: linghaibin@westlake.edu.cn}\\ %
     \scriptsize Westlake University %
\and Bingyao Huang\orcidlink{0000-0002-8647-5730}\thanks{e-mail: bhuang@swu.edu.cn. Corresponding author.}\\ %
     \scriptsize Southwest University}

\teaser{
  \centering
  \includegraphics[width=\linewidth]{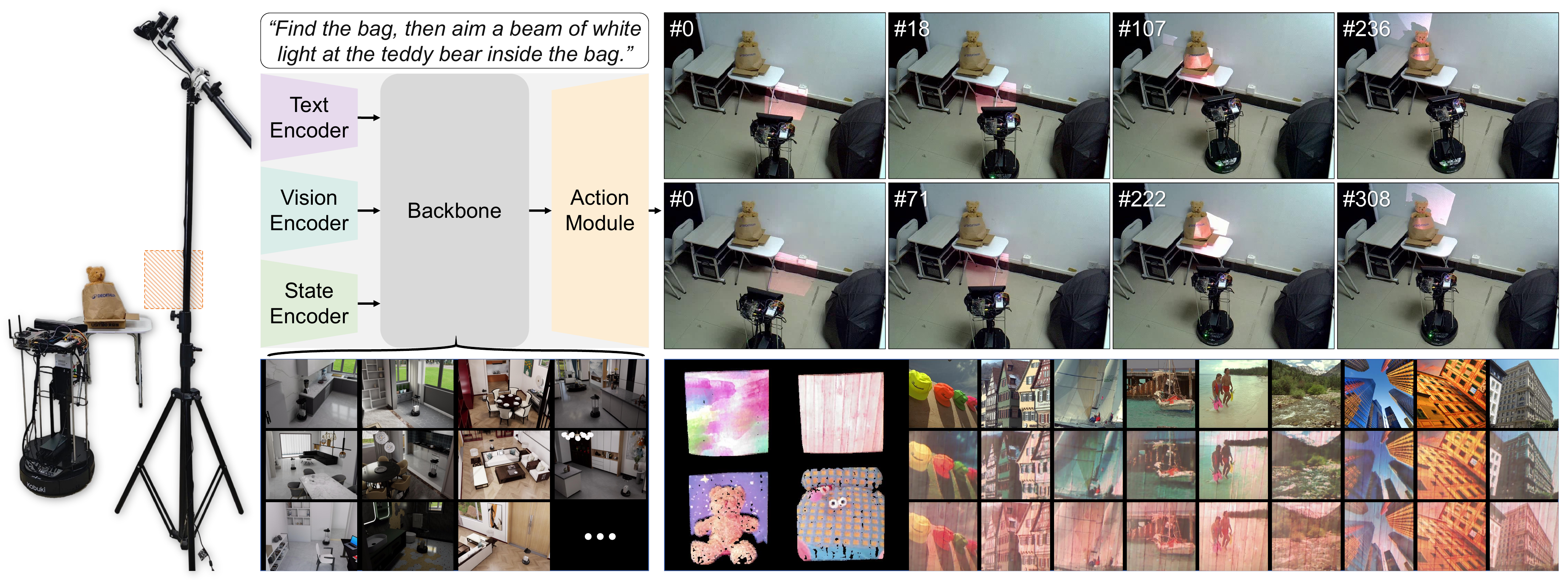}
\caption{\textbf{LICOA makes projected light an embodied action.}
\textit{Left}: The mobile optical embodiment combines a differential-drive chassis for room-scale approach and repositioning with a 2DoF steerable mirror for rapid local beam redirection and projection scale for footprint control. External and onboard RGB cameras provide complementary scene observations.
\textit{Center}: A matched simulation corpus supports adapting pretrained action models through a shared perception and optical action interface, which maps a language instruction, the external and onboard RGB images, and the robot control state to action chunks for the same physical components.
\textit{Upper right}: Selected frames from two real-world rollouts of the displayed instruction, starting from different initial configurations (top and bottom rows); the indices indicate progression within each rollout. Both executions coordinate chassis motion and mirror steering and end with the projection placed on the instructed target. Complete rollouts are shown in the supplementary video.
\textit{Lower right}: Structured light reconstruction and projector compensation results illustrate additional operations after spatial placement.
}
  \label{fig:teaser}
}

\input{sections/00_abstract}

\begin{document}

\maketitle

\input{sections/01_introduction}
\input{sections/02_related_work}
\input{sections/03_method}
\input{sections/04_experiments}
\input{sections/05_discussion}
\input{sections/06_conclusion}

\bibliographystyle{abbrv-doi}

\bibliography{references}

\input{supp/supp.tex}

\end{document}

%% file: preamble.tex
\ifpdf
  \pdfoutput=1\relax                   
  \usepackage{graphicx}                
  \DeclareGraphicsExtensions{.pdf,.png,.jpg,.jpeg} 
\else
  \usepackage{graphicx}                
  \DeclareGraphicsExtensions{.eps}     
\fi
\graphicspath{{figures/}{pictures/}{images/}{./}} 

\usepackage{microtype}                 
\PassOptionsToPackage{warn}{textcomp}  
\usepackage{textcomp}                  
\DeclareMathAlphabet{\mathcal}{OMS}{cmsy}{m}{n} 
\usepackage{times}                     
         
\usepackage{cite}                      
\usepackage{tabu}                      
\usepackage{booktabs}
\usepackage{array,booktabs,tabularx}
\usepackage{float}
\usepackage{longtable,booktabs}
\usepackage[ruled,vlined]{algorithm2e}
\usepackage{multirow}
\usepackage{amsmath}
\usepackage{amsfonts}
\usepackage{mathtools}
\usepackage{xcolor} 
\usepackage{colortbl}

\definecolor{red}{RGB}{255, 0, 0} 
\definecolor{red_darker}{RGB}{204, 64, 64}
\definecolor{green}{RGB}{152, 194, 130}
\definecolor{green_darker}{RGB}{112, 154, 90}
\definecolor{yellow}{RGB}{251, 210, 106}

\graphicspath{{figures/}{pictures/}{images/}{./}}
\usepackage{multirow}
 
\usepackage{pifont}
\usepackage[pagebackref=true,breaklinks=true,colorlinks,bookmarks=false,pagebackref,bookmarks]{hyperref}

\renewcommand{\vec}[1]{\boldsymbol{#1}}

\newcommand{\x}{\vec{x}}

\newcommand{\I}{\mathbf{I}}

%% file: sections/00_abstract.tex
\abstract{
Spatial augmented reality (SAR) augments physical environments with projected content, but most systems assume that the projector pose, target region, or motion path has been fixed or specified. We instead make projected light an embodied action: given a language instruction and current visual observations, a system must decide how to move and where and at what scale to place the projection. We introduce LICOA to realize this capability. Its differential-drive chassis provides room-scale approach and repositioning, while a steerable mirror with two degrees of freedom (2DoF) and projection scale provide rapid local control of the beam. A shared interface maps synchronized external and onboard RGB cameras with robot control state to physical optical actions across pretrained action models. Because instruction-conditioned optical action lacks dedicated training data and a common evaluation protocol, we construct a matched simulation embodiment and a corpus with $150$ tasks and $7{,}500$ rendered episodes, then evaluate models trained only on this corpus in previously unseen real-world environments. Across $480$ complete closed-loop real-world execution trials (rollouts) on $12$ tasks, the best adapted action model achieves a $57.50\%$ success rate and $40.64\%$ success weighted by path length, compared with $8.33\%$ and $5.15\%$ for the same architecture trained from scratch. A further $150$ real-world ablation rollouts support the need for both camera views and coordinated control of the chassis and mirror. These results establish a physical platform, shared perception and optical action interface, corpus, and real-world evaluation protocol for treating projected light as an embodied action.
}

\keywords{Spatial augmented reality, embodied intelligence, optical action, vision-language-action}

%% file: sections/01_introduction.tex
\section{Introduction}
\label{sec:introduction}

Spatial augmented reality (SAR) overlays digital visual content directly onto physical environments using projectors~\cite{bimber2005sar, grundhöfer2018pm, iwai2024pm}. Unlike head-mounted or handheld displays, projection requires no worn or held device, supports shared viewing, and presents spatial cues for navigation, object localization, and collaborative work~\cite{Linder2010luminar,Nazarova2021CobotAR,han2022projectionar,Zhou2026ProjecTA}. Yet useful projection also requires placing light at the intended location,
direction, and scale. Most existing SAR systems rely on fixed projector setups and predefined projection regions. Such configurations provide geometric calibration~\cite{sun2008registration, huang2021a, dong2023adaptive} and support diverse projection mapping applications~\cite{Punpongsanon2015softar, bermano2017makeuplamps, huang2021DeProCams, huang2022CompenNeSt++, erel2023nepmap, erel2024casper, Li2025DPCS, Deng2025GS-ProCams}, but restrict the workspace to surfaces visible from a predetermined projector pose. Mobile and steerable systems expand the reachable workspace, but their targets, paths, or control rules are commonly predefined~\cite{Pinhanez2001everywhere,Kjeldsen2002steerable,Wilson2012Steerable,FernandezRodicio2017projection,Tatsumoto2017tracking,han2022projectionar,Zhou2026ProjecTA,Richter2026projection}. They therefore do not address a general case in which a system enters an unfamiliar environment, interprets a language instruction, and decides both how to move and where to place light. Achieving this capability requires treating the placement of projected light as part of the system's physical action. A language instruction such as ``project onto the bear'' specifies a desired physical effect rather than a target robot pose or a preselected display surface. The system must identify the referenced region, approach a feasible projection pose, redirect the beam, adjust its scale, and determine when placement is complete. 

We call this task \emph{instruction-conditioned optical action}. Throughout, \emph{physical} describes tangible entities and actuator-level quantities, such as scene objects, the platform, and its actions, whereas \emph{real-world} denotes the non-simulated deployment and evaluation domain. Realizing this capability introduces three practical challenges.
First, room-scale coverage and local beam placement require different controls. Moving the projector offers broad reach but is slow and changes the sensing viewpoint, whereas beam steering is fast but locally bounded. Multi-projector installations provide coverage at the cost of fixed infrastructure and calibration~\cite{Cruz-Neira1992cave,Raskar2001shaderlamps,Jones2014RoomAlive}, while arm-mounted systems provide local pose flexibility with a more complex mechanism and a workspace bounded by the arm~\cite{Linder2010luminar,Cabrera2020maskbot,Nazarova2021CobotAR}. A mobile SAR
platform must therefore combine broad repositioning with fast local optical control in one action space.
Second, pretrained action models target embodiments such as manipulators and grippers, with different camera inputs, state dimensions, action outputs, and training objectives~\cite{Brohan2022rt1,ONeill2023openx,Ghosh2024octo,Kim2024openvla,Black2024pi0,PhysicalIntelligence2025pi05}. Training from scratch must learn both general vision-language grounding and optical action semantics from task-specific data. We therefore ask whether pretrained models can retain their grounding and action-generation knowledge while learning optical action from task-specific demonstrations~\cite{Brohan2023rt2,ONeill2023openx,PhysicalIntelligence2025pi05}.
Third, pretraining does not remove the need for task-specific data and evaluation. No existing corpus pairs spatial instructions and external and onboard RGB images with coordinated commands for the differential-drive chassis, a steerable mirror with two degrees of freedom (2DoF), and projection scale, plus episode termination outputs, and no common evaluation protocol assesses their physical outcomes. This absence is the underlying data problem; simulation is our means of addressing it. A useful real-world evaluation protocol must match the physical embodiment and sensing roles and test whether models trained without real-world demonstrations can act in unseen real-world settings.

To address these challenges, we introduce LICOA (Learning Instruction-Conditioned Optical Actions), a mobile SAR system that makes projected light an embodied action, as summarized in~\cref{fig:teaser}. First, its differential-drive chassis provides coarse room-scale approach and repositioning, whereas its 2DoF steerable mirror provides rapid local beam redirection and projection scale controls the footprint. The resulting action components jointly determine where light is placed. Second, LICOA combines timestamp-synchronized external and onboard RGB images with the robot control state in a shared perception and optical action interface. This interface maps the observations and action output of different pretrained action models~\cite{PhysicalIntelligence2025pi05,Pertsch2025fast,yuan2026fastwam} to the same physical embodiment. Third, we reproduce the embodiment in simulation~\cite{NVIDIAIsaacSim,InteriorAgent2025} and construct $150$ spatial placement tasks with $7{,}500$ rendered episodes. Models are fine-tuned only on this corpus and evaluated through real-world rollouts with new scenes, targets, instructions, and initial configurations.
Experiments evaluate three action-generation paradigms, isolate pretraining with a scratch version of $\pi_{0.5}$~\cite{PhysicalIntelligence2025pi05}, and ablate the camera views and action components. Across $480$ complete closed-loop real-world executions (rollouts), adapted $\pi_{0.5}$ achieves $57.50\%$ success and $40.64\%$ success weighted by path length, compared with $8.33\%$ and $5.15\%$ for the scratch control; another $150$ rollouts support the ablations.

Our contributions can be summarized as follows:
\begin{itemize} [topsep=0pt,itemsep=0pt,partopsep=0pt, parsep=0pt]
    \item We formulate projected light as an instruction-conditioned optical action and realize it with a mobile optical platform whose differential-drive chassis, 2DoF steerable mirror, projection scale, and episode termination output form a coupled physical action space.
    \item We define a shared perception and optical action interface comprising external and onboard RGB images, the robot control state, action chunks, and physical action components, and instantiate it with pretrained action models based on continuous action generation, autoregressive action tokenization, and world-action modeling.
    \item We establish a matched simulation corpus and real-world evaluation protocol for this new task, including $150$ spatial placement tasks, $7{,}500$ rendered episodes, fine-tuning only in simulation, a real-world evaluation task set of $12$ previously unseen real-world tasks with $480$ real-world evaluation rollouts, and $150$ ablation rollouts that test the sensing and hardware design.
\end{itemize}

%% file: sections/02_related_work.tex
\section{Related Work}
\label{sec:related}

\input{tables/device_comparison.tex}

\subsection{Spatial augmented reality and mobile projection}
Spatial augmented reality (SAR) overlays digital information directly onto physical surfaces and supports shared viewing without worn or handheld displays~\cite{bimber2005sar, grundhöfer2018pm, iwai2024pm}.
To place an augmentation correctly, an autonomous SAR system must perceive candidate surfaces and semantic targets in the scene.
If projected content is visible in the camera input, the system must also distinguish it from the underlying scene. Most projector-camera systems (ProCams) pipelines focus on geometric and photometric perception under a fixed, calibrated setup~\cite{sun2008registration, huang2021a, dong2023adaptive}. Temporal~\cite{inokuchi1984range, srinivasan1984automated} or spatial~\cite{kawasaki2008d, huang2021a} structured light patterns establish dense correspondences for calibration, shape recovery, and appearance control~\cite{kaminokado2019augment}, while high-speed and infrared capture reduce motion-to-photon error on moving faces and hands~\cite{bermano2017makeuplamps, erel2024casper}. Recently, ProCap extends SAR perception from geometry and photometry to semantics by separating the physical scene from projected content and describing the two layers independently~\cite{Cao2026ProCap}. It therefore exposes the virtual-physical ambiguity relevant to visually guided SAR, but stops at scene description rather than selecting and executing a projection action. Once the scene and projector response are known, a complementary line of work determines what a configured projector should emit. Per-pixel correction and learned visual models compensate for geometric and photometric distortions~\cite{raskar2003iLamps, chen2008colormixing, grundhofer2015robust, huang2017radiometriccompensation, luo2021staypositive, huang2019compennet, huang2019CompenNet++, huang2022CompenNeSt++}. Neural rendering, path tracing, and Gaussian splatting model projector-camera light transport for relighting, compensation, or optical simulation~\cite{huang2021DeProCams, erel2023nepmap, Li2025DPCS, Deng2025GS-ProCams}. Application-specific systems edit perceived softness, faces, glossy objects, or hands~\cite{Punpongsanon2015softar, bermano2017makeuplamps, kaminokado2019augment, erel2024casper}, while neural deblurring and alternative optical paths address defocus, shadowing, and ambient illumination~\cite{Kageyama2020ProDebNet, Kageyama2024deblur, Kusuyama2024deblur, Hiratani2023ShadowlessProjectionMapping, yasui2024pm}. LAPIG further maps language prompts to surface-adapted, stylized content, but still assumes a calibrated projector and a given target surface~\cite{Deng2025LAPIG}. Together, these perception and appearance methods define fixed ProCams represented by the first row of~\cref{tab:embodied_device_comparison}, but they do not expose physical controls for deciding where light should be placed.

To distinguish configured projection from autonomous placement,~\cref{tab:embodied_device_comparison} compares architecture classes along five requirements: projector-origin mobility, exposed action components, spatial instruction, learned closed-loop control, and a matched simulation interface. Fixed multi-projector systems enlarge an installed workspace by coordinating calibrated units, but their origins and coverage remain predetermined~\cite{Cruz-Neira1992cave, Raskar2001shaderlamps, Jones2014RoomAlive, siegl2015dpm, nomoto2022dpm}. Fixed beam-steered systems add rapid local redirection through a rotating mirror or pan-tilt mechanism, yet cannot approach targets outside the useful range of their installed origin~\cite{Pinhanez2001everywhere, Kjeldsen2002steerable, Wilson2012Steerable}. Arm-mounted projectors add local pose flexibility for gesture interaction, target tracking, or expressive movement, while remaining bounded by the arm workspace and mechanism complexity~\cite{Linder2010luminar, Cabrera2020maskbot, Nazarova2021CobotAR, hu2025elegnt}. Mounting a projector on a chassis makes its origin room-scale. With fixed optics, mobile systems detect suitable surfaces or visualize a planned navigation path, but the beam direction remains coupled to the chassis pose~\cite{FernandezRodicio2017projection, han2022projectionar}. Adding pan-tilt or gimbal actuation decouples local beam direction from chassis orientation: prior systems use this combination for marker-based tracking, scripted near-object overlays, or semantic-map and geometry-based surface search~\cite{Tatsumoto2017tracking, Zhou2026ProjecTA, Richter2026projection}. These systems establish the mechanical ingredients of mobile projection, but their spatial targets and motion are supplied by detected planes, markers, planned paths, scripts, maps, or scoring rules. None jointly learns closed-loop optical actions from current RGB observations and a natural-language spatial goal, and the optical simulators used by some fixed ProCams methods do not reproduce the mobile embodiment, camera roles, robot control state, and action interface.

LICOA instead connects a natural-language spatial goal and current observations to learned placement actions for a room-scale chassis, 2DoF steerable mirror, and projection scale. As summarized in~\cref{tab:embodied_device_comparison}, its distinction lies in the learned closed-loop optical action interface and matched embodiment.

\subsection{Generalist action models for optical action}
The hardware comparison in~\cref{tab:embodied_device_comparison} identifies learned closed-loop control as a principal capability missing from prior mobile projection systems. LICOA addresses this gap not by designing another action-model backbone, but by adapting pretrained generalist action models to a new optical embodiment. This adaptation is nontrivial: unlike conventional manipulation, the model must coordinate differential-drive velocities, absolute mirror pan and tilt targets, projection scale, and termination from external and onboard RGB observations and the current robot control state. We therefore use generalist action models to study whether pretrained representations and diverse action-generation mechanisms transfer to instruction-conditioned optical action.

\noindent\textbf{Vision-language-action models.}
Vision-language-action (VLA) models learn mappings from visual observations and language instructions to executable controls. RT-1 and RT-2 established scalable language-conditioned robot policies and the transfer of semantic knowledge from vision-language pretraining, while Open X-Embodiment, Octo, and OpenVLA extended this direction through cross-robot data and open generalist policies~\cite{Brohan2022rt1,Brohan2023rt2,ONeill2023openx,Ghosh2024octo,Kim2024openvla}. For LICOA, two action-generation paradigms are particularly relevant. $\pi_{0.5}$ uses a continuous flow-matching action expert, providing a direct mechanism for generating coupled optical action chunks~\cite{Black2024pi0,PhysicalIntelligence2025pi05}. FAST instead compresses continuous action sequences into discrete tokens, allowing $\pi_0$-FAST to generate action chunks autoregressively~\cite{Pertsch2025fast}. Instantiating both through the same optical action interface tests whether LICOA depends on one particular action representation. Comparing the adapted and scratch versions of $\pi_{0.5}$ additionally isolates whether pretrained visual-language and action representations transfer to optical placement.

\noindent\textbf{World-action models.}
World-action models (WAMs) use visual dynamics as an additional learning signal. Earlier video-based robot models generate or encode possible visual futures and convert them into actions through inverse dynamics or downstream control~\cite{Du2023unipi,Du2023vlp,Zhou2024robodreamer}. More recent models jointly learn video evolution and robot actions, including GR-1, GR-2, Video Prediction Policy, Unified Video Action Model, and Unified World Models~\cite{Wu2023gr1,Cheang2024gr2,Hu2024vpp,Li2025uva,Zhu2025uwm}. Fast-WAM retains this video-action co-training signal but predicts actions without explicitly generating future video at inference~\cite{yuan2026fastwam}. We include it as a third instantiation to test whether a video-derived dynamics prior transfers to optical action while remaining compatible with LICOA's closed-loop replanning interface.

Taken together, $\pi_{0.5}$, $\pi_0$-FAST, and Fast-WAM are not presented as an exhaustive survey or as directly comparable pretrained architectures. They instantiate three action-generation paradigms: continuous flow matching, autoregressive action tokenization, and video-action co-training. LICOA supplies each model with the same instruction, external and onboard RGB images, and robot control state, and maps its predictions to the same chassis, mirror, projection-scale, and termination actions. This controlled interface allows the experiments to test model transfer and interface compatibility across action-generation paradigms without changing the physical task or embodiment.

%% file: tables/device_comparison.tex
\begin{table*}[t]
\centering
\caption{Representative SAR architecture classes evaluated against the requirements of instruction-conditioned optical action. Cells summarize the range demonstrated by the cited systems in each class; an individual system may not support every listed capability. The comparison concerns physical projection placement rather than projected content quality.}
\label{tab:embodied_device_comparison}

\begingroup
\scriptsize
\setlength{\tabcolsep}{1.6pt}
\renewcommand{\arraystretch}{1.18}
\newcommand{\fullcell}[1]{\cellcolor{green!30}\textbf{#1}}
\newcommand{\limitedcell}[1]{\cellcolor{yellow!45}#1}
\newcommand{\nonecell}[1]{\cellcolor{red!28}#1}

\begin{tabularx}{\linewidth}{
  @{}
  >{\raggedright\arraybackslash}p{0.27\linewidth}
 *{5}{>{\centering\arraybackslash}X}
  @{}
}
\toprule
\textbf{System / architecture} &
\textbf{Projector origin mobility} &
\textbf{Action components} &
\textbf{Spatial instruction} &
\textbf{Learned closed-loop control} &
\textbf{Matched simulation} \\
\midrule

Fixed ProCams~\cite{huang2021DeProCams,erel2023nepmap,Li2025DPCS,Deng2025GS-ProCams,Deng2025LAPIG} &
\nonecell{Fixed} &
\nonecell{No} &
\nonecell{Content prompt in some} &
\nonecell{No} &
\limitedcell{Optics simulation in some} \\

Fixed multi-projector systems~\cite{Cruz-Neira1992cave,Raskar2001shaderlamps,Jones2014RoomAlive} &
\nonecell{Fixed} &
\nonecell{No} &
\nonecell{No} &
\nonecell{No} &
\nonecell{No} \\

Fixed beam-steered~\cite{Pinhanez2001everywhere,Kjeldsen2002steerable,Wilson2012Steerable} &
\nonecell{Fixed} &
\limitedcell{Optical steering only} &
\limitedcell{Restricted speech / gesture in some} &
\limitedcell{Rules; no learned action model} &
\nonecell{No} \\

Arm-mounted~\cite{Linder2010luminar,Cabrera2020maskbot,Nazarova2021CobotAR,hu2025elegnt} &
\limitedcell{Arm workspace} &
\limitedcell{Arm only} &
\limitedcell{Speech / gesture / tracking / script} &
\limitedcell{Tracking / script; no learned action model} &
\nonecell{No} \\

Mobile + fixed optics~\cite{FernandezRodicio2017projection,han2022projectionar,Stricker2015bot} &
\fullcell{Room-scale} &
\limitedcell{Chassis; fixed optical axis} &
\limitedcell{Detected surface} &
\limitedcell{Geometry; no learned action model} &
\nonecell{No} \\

Mobile + steerable optics~\cite{Tatsumoto2017tracking,Zhou2026ProjecTA,Richter2026projection,Lee2012bot} &
\fullcell{Room-scale} &
\fullcell{Chassis + PT / gimbal} &
\limitedcell{Marker / path / script / search} &
\limitedcell{Servo / planning / script; no learned action model} &
\nonecell{No} \\

\midrule
\textbf{LICOA (Ours)} &
\fullcell{Room-scale} &
\fullcell{Chassis + 2DoF mirror + projection scale} &
\fullcell{Language spatial instruction} &
\fullcell{Adapted action model} &
\fullcell{Matched embodiment + interfaces} \\

\bottomrule
\end{tabularx}

\vspace{2pt}
\parbox{\linewidth}{\scriptsize
\textbf{Legend:} green, fully supports the LICOA requirement; yellow, limited or specialized support; red, not demonstrated.
\textbf{Definitions:}
\emph{Projector origin mobility} describes how far the physical projector can be repositioned, excluding beam steering from a fixed origin;
\emph{action components} identify the physical controls exposed by each system;
\emph{spatial instruction} specifies where and at what scale to project, not only what visual content to render;
\emph{learned closed-loop control} maps the current RGB images and instruction to physical projection actions and replans during execution (yellow denotes non-learned feedback or planning);
\emph{matched simulation} reproduces the embodiment, RGB images, robot control states, and actions under the shared interface;
\emph{PT}, pan-tilt.}
\endgroup
\end{table*}

%% file: sections/03_method.tex
\section{LICOA Physical Platform}
\label{sec:platform}

LICOA is designed around a real-world requirement: projected light must travel across a room and still support fast local placement. We meet these two scales of motion with complementary action components. The differential-drive chassis changes the projector origin for broad approach and repositioning, while the 2DoF steerable mirror redirects the beam locally without rotating the entire platform. Projection scale changes the illuminated footprint, and an episode termination output marks completion. Together, these controls make the placement of light executable as an embodied action.

\begin{figure}[tbp]
  \centering
  \includegraphics[width=\linewidth]{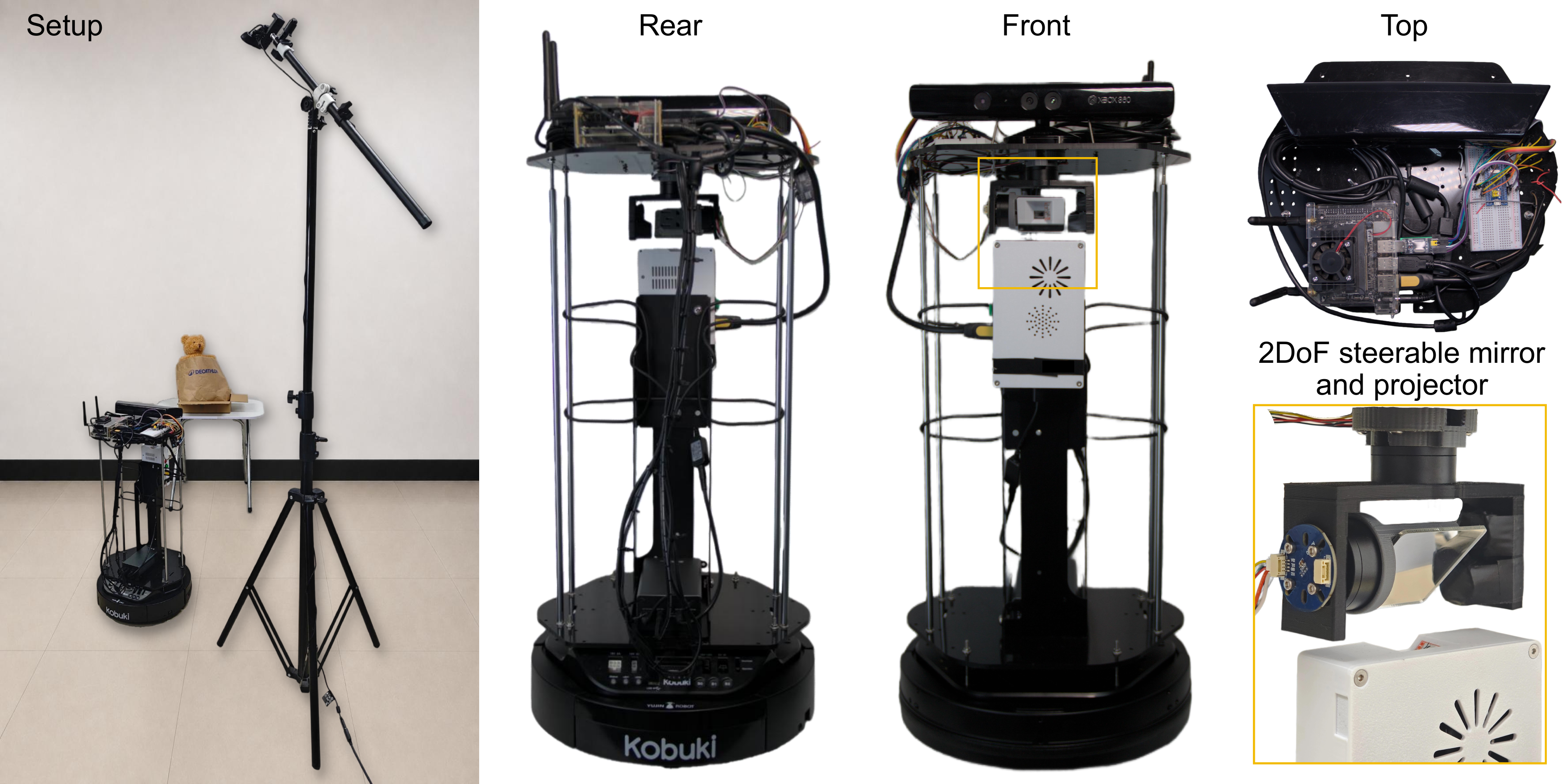}
  \caption{The LICOA physical platform and its complementary motion scales.
  \textit{Setup}: The mobile projection robot and tripod-mounted external RGB camera.
  \textit{Rear}: The Kobuki differential-drive chassis, supporting frame, and integrated electronics for room-scale approach and repositioning.
  \textit{Front}: The onboard RGB camera and projector assembly.
  \textit{Top}: The NVIDIA Jetson Nano and embedded control electronics.
  \textit{2DoF steerable mirror and projector}: The mirror mounted above the projector redirects the beam locally without rotating the differential-drive chassis. Projection scale controls the footprint size.}
  \label{fig:hardware}
\end{figure}

\subsection{Mobile optical embodiment}
\label{subsec:data_acquisition}

\input{tables/hardware}

The hardware shown in~\cref{fig:hardware} and summarized in~\cref{tab:hardware} implements the separation between broad and local control. A rigidly mounted projector would require chassis motion for every change in beam direction; the steerable mirror instead provides pan and tilt while the chassis remains at a suitable pose. Conversely, mirror steering alone covers only a limited local range and cannot reposition the projector. The combined design therefore exposes chassis velocity, mirror pan and tilt, and projection scale as coupled action components. Two onboard battery packs support untethered operation.

The sensing design also serves two spatial roles. The onboard Kinect v1 camera observes the scene from the robot and changes viewpoint as the chassis moves. The external Logitech C270 RGB camera provides a global view of the robot, target region, and projected light. The robot-mounted NVIDIA Jetson Nano acquires the onboard RGB stream and maintains the robot control state accepted by the control stack: differential-drive chassis velocities, pan and tilt targets for the 2DoF steerable mirror, and projection scale. It sends the latest onboard RGB image, timestamp, and robot control state to the host PC as the robot stream, while the external camera stream reaches the host PC directly over USB. The host synchronizes both streams, forms an inference request, receives an action chunk, and relays its per-step actions to the physical action components.

On the physical platform, the Jetson dispatches differential-drive commands to the Kobuki, projection scale commands to the custom laser projector over HDMI, and absolute mirror pan and tilt targets to an STM32 over a universal asynchronous receiver-transmitter (UART) link via USB at $50\,\mathrm{Hz}$ per axis. The STM32 closes both pan-tilt motor loops locally at approximately $260\,\mathrm{Hz}$ using AS5600 encoder feedback. This local loop separates high-rate mirror control from action model inference. Detailed interfaces, calibration, power distribution, and motor control are provided in the Suppl.\ Material.

\section{Shared Perception and Optical Action Interface}
\label{sec:method}

\begin{figure*}
\centering
\includegraphics[width=\linewidth]{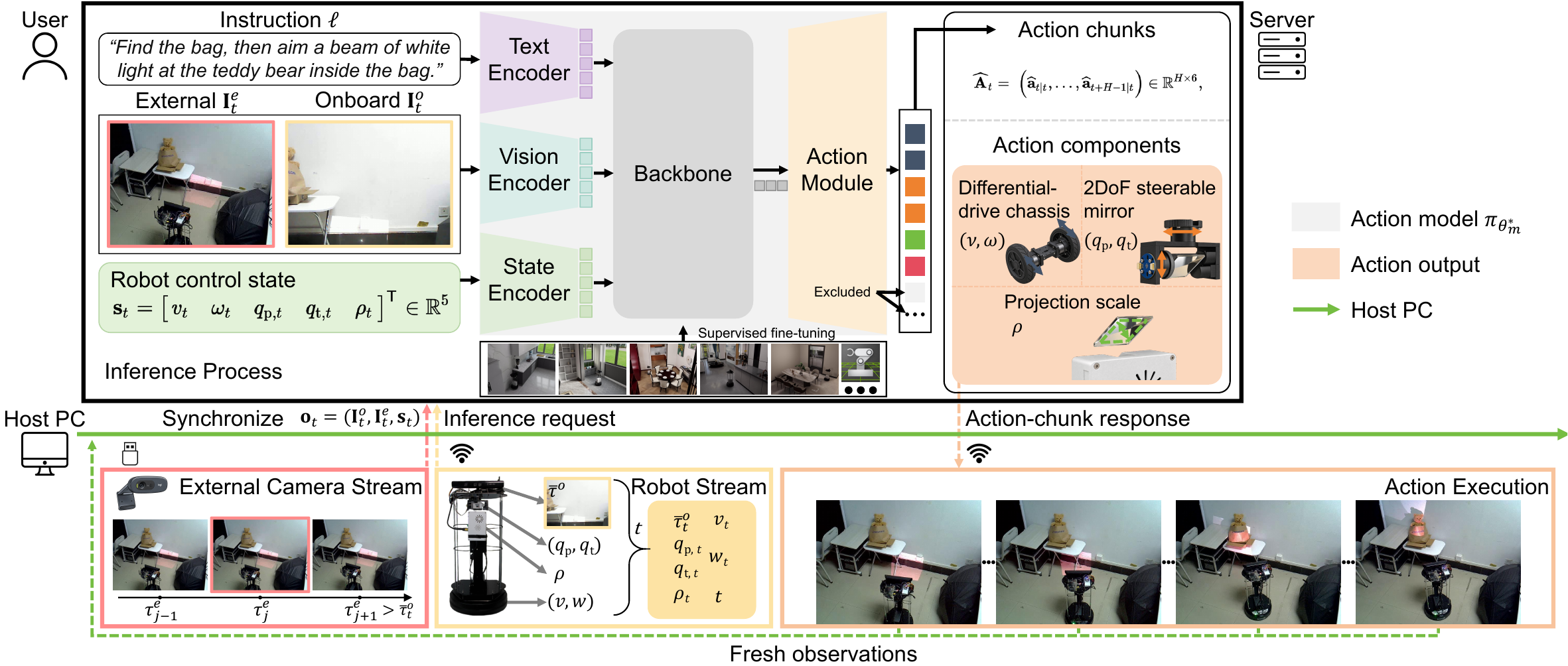}
\caption{LICOA makes projected light executable through a shared perception and optical action interface. \textit{Top}: The instruction $\ell$, external and onboard RGB images, and five-dimensional robot control state pass through the encoders, backbone, and action module. Supervised fine-tuning adapts the action model to return $H$-step action chunks. Each six-dimensional action contains commands for the differential-drive chassis, 2DoF steerable mirror, and projection scale, plus an episode termination output; any additional model output dimensions are excluded. \textit{Bottom}: The host PC synchronizes the external camera stream and robot stream, sends an inference request, executes a short prefix of the returned action chunk, and replans from fresh observations. Simulation training and real-world deployment use the same observation and action interfaces.}
\label{fig:pipeline}
\end{figure*}

The platform becomes learnable through a shared interface that separates physical meaning from model implementation. Every action model receives the same instruction, external and onboard RGB images, and robot control state, then returns action chunks comprising commands for the differential-drive chassis, 2DoF steerable mirror, and projection scale, plus an episode termination output.~\Cref{fig:pipeline} summarizes this interface and its real-world execution loop.

\subsection{Observation and action interface}
\label{sec:problem_formulation}

At each replanning step $t$, the action model must combine local and global visual information. The host PC synchronizes the external camera stream with the robot stream to form an observation comprising the onboard image $\I_t^o$, external image $\I_t^e$, and robot control state $\mathbf{s}_t$, then sends it with instruction $\ell$ as an inference request. Each robot stream packet carries the latest onboard RGB image, its capture timestamp, and the robot control state, while the host continuously buffers timestamped images from the external camera stream. Let $\bar{\tau}_t^o$ be the selected onboard image's capture timestamp expressed on the host clock through periodic clock-offset estimation, and let $\tau_j^e$ be the timestamp of buffered external image $j$. The host selects the latest external image not newer than the onboard image:
\begin{equation}
  j_t=\max\!\left\{j\mid \tau_j^e\leq\bar{\tau}_t^o\right\}.
  \label{eq:timestamp_pairing}
\end{equation}

The selected images are denoted by $\I_t^o$ and $\I_t^e$. The robot control state $\mathbf{s}_t$ is carried by the same robot stream packet as $\I_t^o$. Only the assembled observation at a replanning boundary is sent to the server; intermediate images from the raw camera streams are not individually forwarded. We write the robot control state as
\begin{equation}
  \mathbf{s}_t
  =\begin{bmatrix}
    v_t & \omega_t & q_{\mathrm{p},t} & q_{\mathrm{t},t} & \rho_t
  \end{bmatrix}^{\mathsf{T}}\in\mathbb{R}^{5}.
  \label{eq:control_state}
\end{equation}

Together with the external and onboard RGB images, it forms the timestamp-synchronized observation
\begin{equation}
  \mathbf{o}_t=\left(\I_t^{o},\I_t^{e},\mathbf{s}_t\right).
  \label{eq:observation}
\end{equation}

The shared index $t$ therefore identifies one assembled replanning observation, rather than every image in either raw camera stream. Here, $v_t$ and $\omega_t$ are the active body-frame linear and angular velocities of the differential-drive chassis in $\mathrm{m/s}$ and $\mathrm{rad/s}$, $q_{\mathrm{p},t}$ and $q_{\mathrm{t},t}$ are the absolute mirror pan and tilt targets of the 2DoF steerable mirror in radians, and $\rho_t\in[0,100]$ is the projection scale expressed as a percentage. Thus, $\mathbf{s}_t$ is a synchronized robot control state rather than a complete physical robot state; in particular, it does not include the global chassis pose. The instruction $\ell$ conditions the action model separately because it is a task specification rather than a synchronized sensor or control measurement. The real-world and simulated platforms expose this same observation interface.

Given $(\mathbf{o}_t,\ell)$, the action model predicts an $H$-step action chunk. We separate the continuous physical control from the binary termination decision. For future offset $i$, the continuous control is
\begin{equation}
  \mathbf{u}_{t+i\mid t}
  =\begin{bmatrix}
    v & \omega & q_{\mathrm{p}} & q_{\mathrm{t}} & \rho
  \end{bmatrix}_{t+i\mid t}^{\mathsf{T}}\in\mathbb{R}^{5}.
  \label{eq:continuous_control}
\end{equation}

Appending the binary termination indicator gives the complete per-step action
\begin{equation}
  \mathbf{a}_{t+i\mid t}
  =\begin{bmatrix}
    \mathbf{u}_{t+i\mid t}^{\mathsf{T}} & b_{t+i\mid t}
  \end{bmatrix}^{\mathsf{T}}\in\mathbb{R}^{6}.
  \label{eq:action}
\end{equation}

Stacking $H$ successive actions gives the action chunk
\begin{equation}
  \mathbf{A}_t
  =\left(\mathbf{a}_{t\mid t},\ldots,
  \mathbf{a}_{t+H-1\mid t}\right)\in\mathbb{R}^{H\times 6}.
  \label{eq:action_chunk}
\end{equation}

At the canonical observation and action interface, action model $\pi_{\theta_m}$ induces the conditional distribution
\begin{equation}
  \widehat{\mathbf{A}}_t
  \sim\pi_{\theta_m}\!\left(\cdot\mid\mathbf{o}_t,\ell\right).
  \label{eq:policy_interface}
\end{equation}

Specifically, a sampled prediction is the ordered sequence
\begin{equation}
  \begin{aligned}
    \widehat{\mathbf{A}}_t
    &=\left(\widehat{\mathbf{a}}_{t\mid t},\ldots,
      \widehat{\mathbf{a}}_{t+H-1\mid t}\right)
      \in\mathbb{R}^{H\times 6},\\
    \widehat{\mathbf{a}}_{t+i\mid t}
    &\in\mathbb{R}^{6},\qquad 0\leq i<H.
  \end{aligned}
  \label{eq:predicted_action_chunk}
\end{equation}

Within $\mathbf{u}_{t+i\mid t}$, the first two entries are body-frame velocity commands for the differential-drive chassis, and the remaining three are absolute mirror pan and tilt targets for the 2DoF steerable mirror and the projection scale target. The binary indicator $b_{t+i\mid t}\in\{0,1\}$ requests episode termination when it equals one. During learning, this sixth target channel is stored as zero or one. At deployment, the decoded scalar $\widehat{b}_{t+i\mid t}\in\mathbb{R}$ is converted to the termination decision by the controller threshold $\widehat{b}_{t+i\mid t}>0.5$. Although $\mathbf{s}_t$ and $\mathbf{u}_{t+i\mid t}$ use the same five physical coordinates, the former records the current synchronized control values, whereas the latter specifies future controls. This canonical observation and action interface is independent of how action model family $m$ internally represents or generates an action.

\subsection{Connecting pretrained action models}
\label{sec:policy_adaptation}

Different pretrained action models expect different cameras, image layouts, state dimensions, action representations, and training targets. These implementation differences should not change the SAR task. We therefore keep the observation and action interfaces in~\cref{eq:observation,eq:action} fixed and place model-specific conversion only at their boundaries. At each replanning boundary, an input adapter maps the instruction, external and onboard RGB images, and robot control state to the camera slots and preprocessing expected by a model. A target adapter converts the demonstrated physical action into that model's training representation, and an output adapter converts its prediction back to the LICOA action components.

Let $\xi_t$ be a synchronized demonstration segment beginning at $t$. It contains $(\mathbf{o}_t,\ell)$, the target physical action chunk $\mathbf{A}_t^*$, and any aligned future visual observations required by a model's native auxiliary objective. Let $\mathcal{D}$ denote the collection of these demonstration segments. For action model family $m$, the input adapter produces
\begin{equation}
  x_t^{(m)}=\Phi_m^{\mathrm{in}}\!\left(\mathbf{o}_t,\ell\right).
  \label{eq:input_adapter}
\end{equation}
The target adapter converts the demonstration supervision into the corresponding target in the model's native representation
\begin{equation}
  y_t^{(m)}=\Phi_m^{\mathrm{tgt}}\!\left(\xi_t\right).
  \label{eq:target_adapter}
\end{equation}

Let $\mathcal{L}_m$ denote the model's native training objective, and let $\mathcal{U}_m$ index its selected trainable modules, with $\overline{\mathcal{U}}_m$ denoting the complement. We minimize this objective over the selected modules:
\begin{equation}
  \theta_{m,\mathcal{U}_m}^{*}
  = \arg\min_{\theta_{m,\mathcal{U}_m}}
  \mathbb{E}_{\xi_t\sim\mathcal{D}}
  \left[\mathcal{L}_{m}\!\left(
  \theta_m;x_t^{(m)},y_t^{(m)}\right)\right].
  \label{eq:adaptation}
\end{equation}
Here, $\Phi_m^{\mathrm{in}}$ maps the common observation to the model's native input, including camera key mapping, resizing, normalization, and required state padding. The target adapter $\Phi_m^{\mathrm{tgt}}$ expresses the same physical supervision in the model's native form. It may return a normalized continuous chunk, an autoregressive sequence of action tokens, or a joint visual and action target. Accordingly, $\mathcal{L}_m$ remains the native objective of the pretrained action model family: it can incorporate noisy actions and a flow time variable, preceding action tokens, or an auxiliary loss on future visual observations without forcing all architectures to share one surrogate objective. In~\cref{eq:adaptation}, $\theta_m$ comprises the optimized parameters $\theta_{m,\mathcal{U}_m}$ and a complementary set fixed at its pretrained value $\theta^0_{m,\overline{\mathcal{U}}_m}$. The minimizer $\theta_{m,\mathcal{U}_m}^{*}$, together with this frozen complement, defines the full adapted parameters $\theta_m^*$. This covers both full fine-tuning and partial post-training tailored to each model. Regardless of these internal choices, an output adapter maps every prediction back to the canonical physical action space in~\cref{eq:action}. Specific adapter and optimization settings are reported with the evaluated models in~\cref{sec:experimental_setup} and the Suppl.\ Material.

\subsection{Closed-loop action execution}
\label{sec:closed_loop_control}

During deployment, the action model repeatedly observes the current state of the projection task rather than following a fixed trajectory. The timestamp-synchronized external and onboard RGB images, robot control state, and instruction are processed by the same input adapter used during training. Let $p_{\theta_m^*}^{(m)}(\cdot\mid x_t^{(m)})$ denote the predictive distribution induced by the adapted model in its native space; it covers both continuous sampling and autoregressive generation. The adapted action model first produces a prediction in that space
\begin{equation}
  \widehat{y}_t^{(m)}
  \sim p_{\theta_m^*}^{(m)}\!\left(
    \cdot\mid x_t^{(m)}\right).
  \label{eq:model_prediction}
\end{equation}

The output adapter decodes it into physical actions
\begin{equation}
  \widehat{\mathbf{A}}_t
  =\Phi_m^{\mathrm{out}}\!\left(
    \widehat{y}_t^{(m)}\right).
  \label{eq:output_adapter}
\end{equation}

At the canonical observation and action interface, this yields the adapted action model
\begin{equation}
  \widehat{\mathbf{A}}_t
  \sim\pi_{\theta_m^*}\!\left(
    \cdot\mid\mathbf{o}_t,\ell\right).
  \label{eq:adapted_policy}
\end{equation}

Thus,~\cref{eq:model_prediction,eq:output_adapter} give the realization specific
to each model family of~\cref{eq:adapted_policy}, which is the adapted instance of the
generic physical action model in~\cref{eq:policy_interface}.

For $1\leq h\leq H$, the controller dispatches
\begin{equation}
  \widehat{\mathbf{A}}_t^{(h)}
  =\left(\widehat{\mathbf{a}}_{t\mid t},\ldots,
    \widehat{\mathbf{a}}_{t+h-1\mid t}\right)
  \in\mathbb{R}^{h\times 6}.
  \label{eq:execution_chunk}
\end{equation}

Many action chunk models maintain an internal queue so that successive calls can return actions from the same prediction. This is efficient during open-loop execution but can accidentally reuse a chunk generated from an old observation. We hence clear the queue at the beginning of every inference request, which becomes a new replanning boundary: the action model consumes the latest observation, its output adapter maps the prediction back to the six-dimensional LICOA action interface, and the controller executes $h=8$ actions before requesting a new chunk. Repeating this procedure at $15\,\mathrm{Hz}$ yields receding-horizon closed-loop action execution.

\section{Simulation Corpus and Evaluation Protocol}
\label{sec:dataset}

Instruction-conditioned optical action has no established corpus that combines language, two camera roles, robot control state, and mobile optical action components. It also lacks a common task and evaluation protocol. We address both gaps with a matched simulation corpus for training and a real-world evaluation protocol.

\subsection{Matched simulation and data collection}

We reproduce the physical LICOA embodiment in NVIDIA Isaac Sim~\cite{NVIDIAIsaacSim} using InteriorAgent USD indoor scenes~\cite{InteriorAgent2025}. The simulation matches the physical kinematic chain, camera configuration, action ranges, and projection scale so that both embodiments share the observation and action interfaces defined in~\cref{sec:problem_formulation}.

We collect expert demonstrations in simulation by teleoperating the differential-drive chassis, 2DoF steerable mirror, and projection scale according to a language instruction. Each accepted demonstration forms a source trajectory that starts from an initial chassis and mirror configuration and ends when the instructed projection is achieved or the maximum trajectory length is reached. Repeated source trajectories vary the initial configuration and target specification, producing both short local corrections and trajectories that require chassis repositioning before mirror adjustment. At each time step, we record the instruction, timestamp-synchronized external and onboard RGB images, the five-dimensional robot control state, and the corresponding six-dimensional action defined in~\cref{eq:control_state,eq:action}. The physical and simulated platforms serialize these observations and actions in the same trajectory format.

\subsection{Task collection}
\label{subsec:dataset_composition}

\input{tables/dataset}

The matched simulation corpus defines spatial placement tasks rather than content generation ones. We select the first $10$ InteriorAgent indoor scenes, each a complete environment rather than an individual room or camera view. Within each scene, we define $5$ fixed external camera viewpoints and construct three tasks per viewpoint, yielding $15$ tasks per scene and $150$ tasks in total. A task is specified by its scene and external camera viewpoint, an initial robot configuration, and a language instruction describing a target surface or object-relative region and the intended projection scale. Examples include ``project near the chair,'' ``move the light to the left of the table,'' and ``place a larger light patch on the wall above the cabinet.''

For the $150$ tasks, we collect $10$ source trajectories per task and render each trajectory with $5$ appearance settings: the original appearance and four variants that randomize lighting intensity, lighting color temperature, wall materials, and floor materials. This procedure yields $50$ rendered episodes per task and $7{,}500$ rendered episodes in total while preserving the underlying geometry and source action sequence. Each episode includes the language instruction, timestamp-synchronized external and onboard images, robot control states, and actions; geometric target masks, projection centroids, and projection scales are not stored. \Cref{tab:dataset} summarizes the corpus. All $10$ scenes contribute training data for both main experiments and ablations, and configurations using the corpus include all $150$ tasks. The source trajectory holdout is used only for offline validation, not as a benchmark for unseen scenes or instructions, as described in Suppl.\ Material. Instead, the real-world evaluation protocol assesses deployment in unseen environments with new instructions that do not appear in the matched simulation corpus.

\subsection{Real-world evaluation protocol}
\label{subsec:body_optical_protocol}

The real-world evaluation protocol measures the final physical placement of a uniform white rectangular projection rather than only the robot pose. A correct chassis pose is insufficient if the mirror or projection scale leaves the beam outside the instructed region. Because the corpus does not store geometric target masks or calibrated target centroids and scales, we do not apply an automatic geometric success metric. Instead, a human operator judges whether the final projection satisfies the spatial instruction. A rollout is counted as successful when the operator confirms the placement and either the decoded episode termination output triggers termination or the projection remains stable on the instructed target for at least $3\,\mathrm{s}$. The resulting annotation is used for real-world metrics (\cref{sec:experiments}).

%% file: tables/hardware.tex

\begin{table}[t]
  \centering
  \caption{Hardware roles in the physical LICOA platform. The chassis provides broad repositioning, while the mirror and projector provide local beam direction and scale control. Detailed interfaces, electronics, power distribution, and embedded control are provided in the Suppl.\ Material.}
  \label{tab:hardware}
  \footnotesize
  \renewcommand{\arraystretch}{1.12}
  \begin{tabularx}{\columnwidth}{@{}
    >{\raggedright\arraybackslash}p{0.22\columnwidth}
    >{\raggedright\arraybackslash}p{0.31\columnwidth}
    >{\raggedright\arraybackslash}X@{}}
    \toprule
    \textbf{Subsystem} & \textbf{Hardware} & \textbf{Role} \\
    \midrule
    Differential-drive chassis
      & Kobuki mobile chassis
      & Room-scale approach and repositioning \\
    Onboard sensing
      & Microsoft Kinect v1
      & Robot-centric RGB observation \\
    External sensing
      & Logitech C270
      & Global RGB observation \\
    Embedded computing
      & NVIDIA Jetson Nano
      & Robot-side sensing and execution bridge \\
    Projection
      & Custom laser projector
      & Optical output and footprint scaling \\
    Optical steering
      & Custom 2DoF steerable mirror
      & Rapid local beam redirection \\
    Pan-tilt servo control
      & STM32F103C8T6
      & Low-level motor feedback control \\
    \bottomrule
  \end{tabularx}
\end{table}

%% file: tables/dataset.tex

\begin{table}[t]
  \centering
  \caption{Composition of the LICOA matched simulation corpus. We use the first $10$ InteriorAgent scenes, with five external camera viewpoints and three spatial placement tasks per viewpoint, yielding $150$ tasks. Each source trajectory is rendered with the original appearance and four randomized variants in lighting, wall, and floor materials. Trajectory density and task diversity analyses are provided in the Suppl.\ Material.}
  \label{tab:dataset}
  \resizebox{\columnwidth}{!}{%
    \begin{tabular}{@{}cccc@{}}
    \toprule
    \shortstack{\textbf{Scenes}\\$10$} &
    \shortstack{\textbf{External viewpoints/scene}\\$5$} &
    \shortstack{\textbf{Tasks/external viewpoint}\\$3$} &
    \shortstack{\textbf{Total tasks}\\$150$} \\
    \midrule
    \shortstack{\textbf{Source trajectories/task}\\$10$} &
    \shortstack{\textbf{Appearances/trajectory}\\$5$} &
    \shortstack{\textbf{Total episodes}\\$7{,}500$} &
    \\
    \bottomrule
    \end{tabular}
  }

  \includegraphics[width=\linewidth]{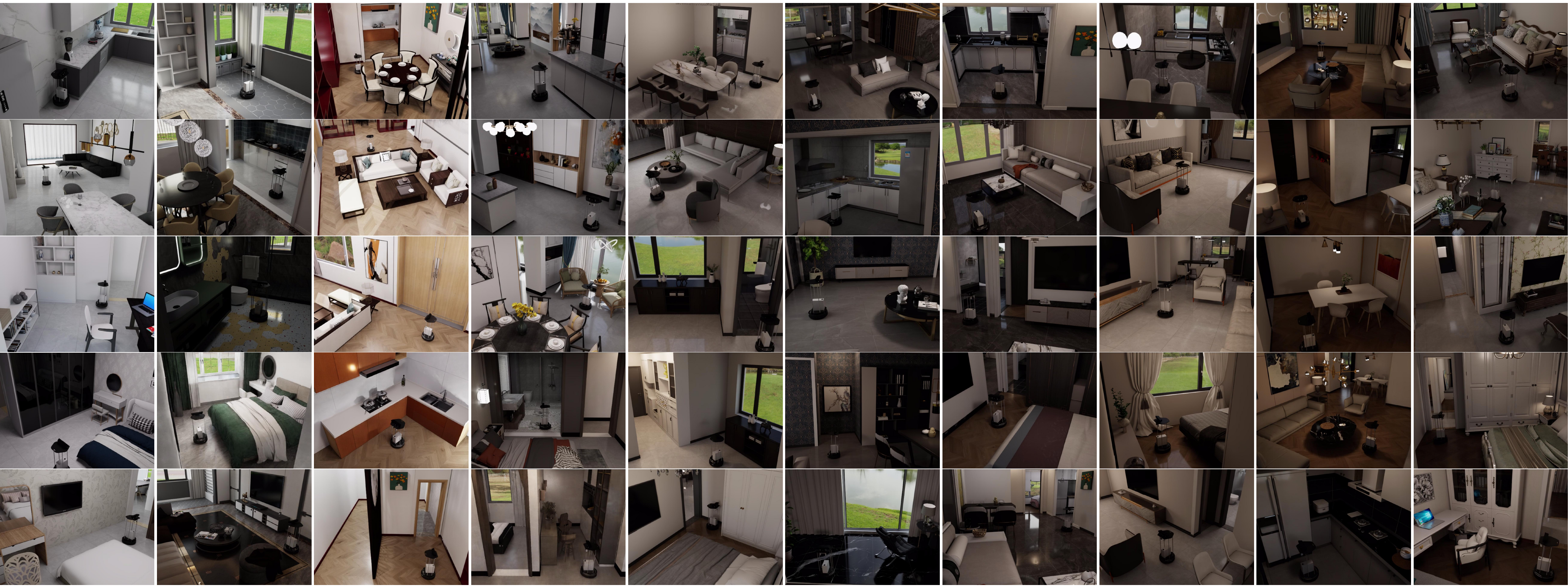}
\end{table}

%% file: sections/04_experiments.tex
\section{Experiments}
\label{sec:experiments}

The experiments evaluate the three contributions in~\cref{sec:introduction}: three pretrained model families test interface generality; simulation-only training followed by real-world deployment in unseen environments, together with a $\pi_{0.5}$-scratch control, assesses transfer and the value of pretraining; and camera and action component ablations examine the complementary roles of the external and onboard sensing, chassis motion, and mirror steering.

\subsection{Experimental setup}
\label{sec:experimental_setup}

\noindent\textbf{Action models and pretraining.}
To test the shared interface in~\cref{sec:policy_adaptation}, we instantiate the optimization in~\cref{eq:adaptation} with three action model families spanning continuous action generation, autoregressive action tokenization, and world-action modeling. The comparison includes $\pi_{0.5}$, a continuous generative VLA model~\cite{PhysicalIntelligence2025pi05}; $\pi_0$-FAST, which autoregressively generates tokenized actions~\cite{Pertsch2025fast}; and Fast-WAM~\cite{yuan2026fastwam}, which builds on video-generative representations. We also train a $\pi_{0.5}$-scratch control with the same architecture and LICOA training data but without pretrained initialization. This control isolates the contribution of the representations acquired during large-scale pretraining.

\noindent\textbf{Common simulation training and real-world deployment.}
All methods use the same demonstrations, external and onboard RGB images, five-dimensional robot control state, six-dimensional action space, and $16$-step action horizon defined by the interfaces in~\cref{eq:observation,eq:action}. Each model uses the family-specific adapters in~\cref{eq:input_adapter,eq:target_adapter,eq:output_adapter}, namely $\Phi_m^{\mathrm{in}}$, $\Phi_m^{\mathrm{tgt}}$, and $\Phi_m^{\mathrm{out}}$, while retaining its native action representation and training objective. On the platform, the external and onboard RGB cameras are configured to acquire $640\times480$ images at $30\,\mathrm{Hz}$; the host applies the timestamp pairing rule in~\cref{eq:timestamp_pairing} before each inference request. Each run uses $30{,}000$ optimizer updates, batch size $32$, and one NVIDIA H200 GPU. We retain each model's native optimizer and learning rate schedule. The pretrained $\pi_{0.5}$ and $\pi_0$-FAST models are fully fine-tuned without low-rank adaptation (LoRA) or other parameter-efficient fine-tuning (PEFT) modules; Fast-WAM jointly trains its video and action experts. The scratch uses the same full $\pi_{0.5}$ recipe without loading a pretrained checkpoint. During real-world evaluation, the server returns a $16$-step action chunk at each replanning request. The controller dispatches its first eight per-step actions at a nominal $15\,\mathrm{Hz}$ before replanning from fresh observations. Model-specific optimizer and scheduler settings are provided in the Suppl.\ Material.

\noindent\textbf{Real-world evaluation task set.}
Following the success criterion defined in~\cref{subsec:body_optical_protocol}, we evaluate each action model on $12$ tasks in previously unseen real-world environments. The tasks contain new spatial instructions absent from the matched simulation corpus, together with different target objects, initial robot configurations, and required optical actions. Each model is evaluated for $10$ rollouts per task, yielding $120$ rollouts per model and $480$ rollouts in total. We report success rate (SR), success weighted by path length (SPL), and mean completion time (MT). For rollout $i$, let $S_i\in\{0,1\}$ be its success indicator, $L_i$ the straight-line displacement between its initial and final chassis positions, $P_i$ its executed chassis path length, and $T_i$ its wall-clock execution time. We first define path efficiency $e_i$ with $\varepsilon=10^{-4}\,\mathrm{m}$ to cover the case with no chassis displacement:
\begin{equation}
  e_i=
  \begin{cases}
    1, & L_i\leq\varepsilon\ \text{and}\ P_i\leq\varepsilon,\\
     {L_i}\big/{\max\!\{L_i,P_i\}},
     & \text{otherwise}.
  \end{cases}
  \label{eq:path_efficiency}
\end{equation}
Then, for $N$ rollouts, we define
\begin{equation}
  \mathrm{SR}\!=\!\frac{1}{N}\sum_{i=1}^{N}S_i, \ \
  \mathrm{SPL}\!=\!\frac{1}{N}\sum_{i=1}^{N}S_i e_i, \ \
  \mathrm{MT} \!=\!\frac{\sum_{i=1}^{N}S_iT_i}{\sum_{i=1}^{N}S_i}.
  \nonumber
\end{equation}

SPL penalizes both failure and inefficient motion of the differential-drive chassis, whereas MT measures execution speed only among successful trials. For real-world rollouts, $L_i$ and $P_i$ are computed by midpoint integration of the linear and angular velocities measured by the Kobuki chassis. Since an optical task can admit multiple valid terminal chassis poses, $L_i$ is measured to the achieved successful endpoint. SR and SPL are computed over all $120$ rollouts per action model, while MT is computed over the successful subset.

\subsection{Interface generality and real-world deployment}
\label{sec:real_world_results}

\begin{figure*}[t]
  \centering
  \includegraphics[width=\linewidth]{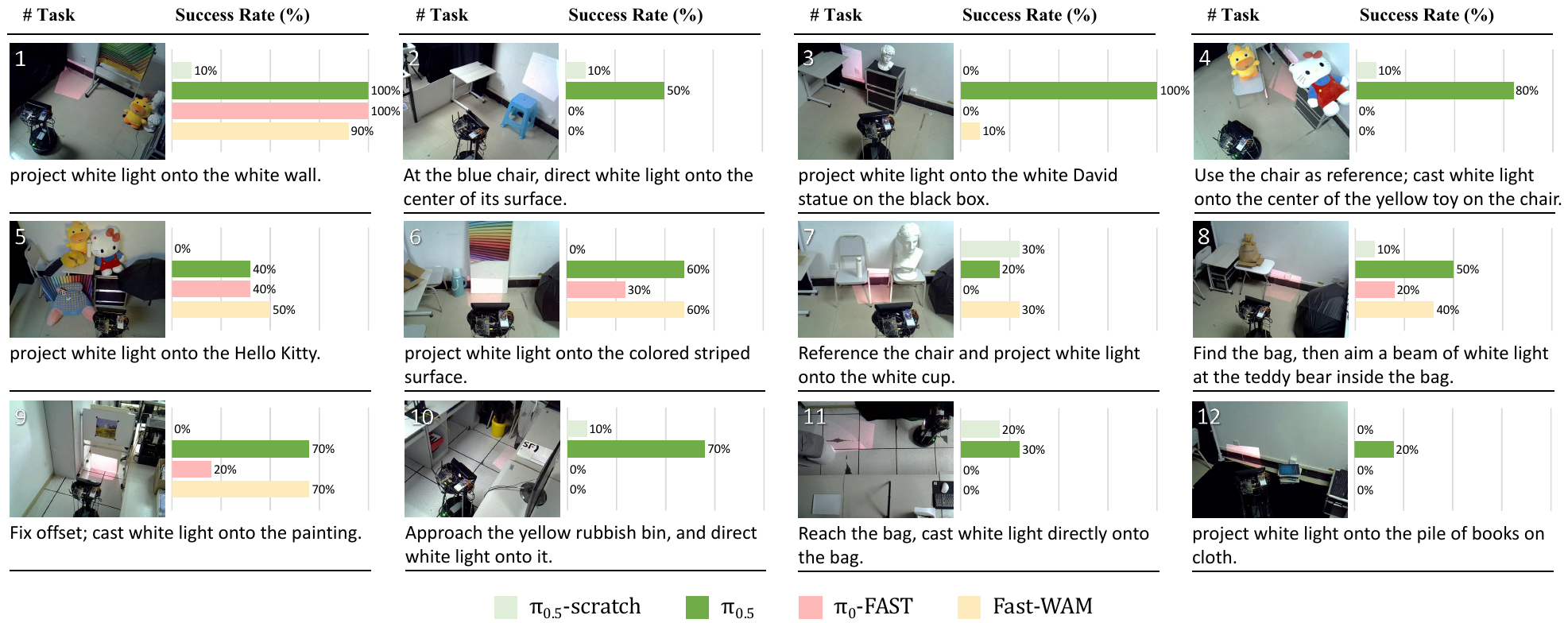}
  \caption{Per-task evidence for real-world deployment through the shared perception and optical action interface. Each action model is trained only on the matched simulation corpus and evaluated for $10$ rollouts on each of $12$ previously unseen real-world tasks. The instructions and target scenes are shown in reading order, and the bars report success rate. All tasks contribute equally to the overall SR in~\cref{tab:main_comparison}.}
  \label{fig:qualitative_exp}
\end{figure*}

\input{tables/comparison}

Aggregate comparisons and per-task results are shown in~\cref{tab:main_comparison} and~\cref{fig:qualitative_exp}, respectively. All three pretrained action model families achieve nonzero success after being connected to LICOA through the same physical interface and trained only in simulation. This result demonstrates that the interface can support continuous generation, autoregressive action tokenization, and world-action modeling, although their real-world performance differs substantially. Adapted $\pi_{0.5}$ achieves the highest overall SR of $57.50\%$ ($69/120$), outperforming Fast-WAM, $\pi_0$-FAST, and $\pi_{0.5}$-scratch by $28.33$, $40.00$, and $49.17$ percentage points, respectively. It is also the only action model to succeed at least once on every task. In comparison, Fast-WAM, $\pi_0$-FAST, and $\pi_{0.5}$-scratch obtain nonzero success on $7$, $5$, and $7$ of the $12$ tasks, respectively. These results show that the advantage of $\pi_{0.5}$ is not restricted to a small subset of targets. The same model achieves the best SPL ($40.64\%$) and the lowest MT over successful trials ($43.44\,\mathrm{s}$). Fast-WAM ranks second in both SR and SPL, but its SPL of $21.76\%$ remains $18.88$ percentage points below that of $\pi_{0.5}$. Although $\pi_0$-FAST produces relatively direct chassis trajectories when it succeeds, its low success frequency reduces its overall SPL to $15.81\%$, and its successful rollouts require $139.38\,\mathrm{s}$ on average. The low MT of $\pi_{0.5}$-scratch should be interpreted together with its $8.33\%$ SR: because MT is conditioned on success, it reflects only the few tasks completed by this baseline and does not indicate robust execution.

\noindent\textbf{Evidence from pretraining.}
The gap between $\pi_{0.5}$ and its scratch counterpart is particularly large: pretrained initialization improves average SR from $8.33\%$ to $57.50\%$ and SPL from $5.15\%$ to $40.64\%$. Since the two variants share the action model architecture, training data, and action interface, this comparison indicates that pretrained vision-language and action representations are important for adapting to the new optical action embodiment. Nevertheless, the remaining $42.50\%$ failure rate also shows that reliable coordination of target grounding, the differential-drive chassis, and the 2DoF steerable mirror remains challenging.

\subsection{Evidence for sensing and action design}
\label{sec:ablation_studies}

We use the best-performing $\pi_{0.5}$ in~\cref{tab:main_comparison} to ablate the complementary camera views and coupled chassis-mirror control introduced in~\cref{sec:platform}. Camera variants are retrained under the same fine-tuning recipe, with one view removed from both the dataset and model input mapping; action variants retain the full checkpoint and suppress the indicated commands only at deployment. All models train on the same $10$ simulation scenes. Evaluation uses three novel real-world tasks with $10$ rollouts per configuration and task, with instructions and initial configurations matched across configurations.

\noindent\textbf{External and onboard views.}
The full action model receives timestamp-synchronized external and onboard images. We retrain two variants, each removing one view, to test the contribution of global scene context or robot-centric feedback, respectively. Retraining avoids the additional distribution shift that would arise from masking a view only at inference.

\noindent\textbf{Coupled action components.}
We further compare the full coupled action space with two deployment-time restrictions. \emph{Mirror only} replaces both predicted commands for the differential-drive chassis with zero while retaining the commands for the 2DoF steerable mirror and projection scale. \emph{Chassis only} replaces the two predicted pan and tilt targets for the 2DoF steerable mirror with the canonical fixed pose while retaining the commands for the differential-drive chassis and projection scale. Thus, both variants use the same full action model checkpoint and differ only in the commands dispatched to the physical actuators. We report SR, SPL, and MT over successful trials using the same
definitions as in~\cref{sec:experimental_setup}.

\input{tables/ablation}

\begin{figure*}[!t]
  \centering
  \includegraphics[width=\textwidth]{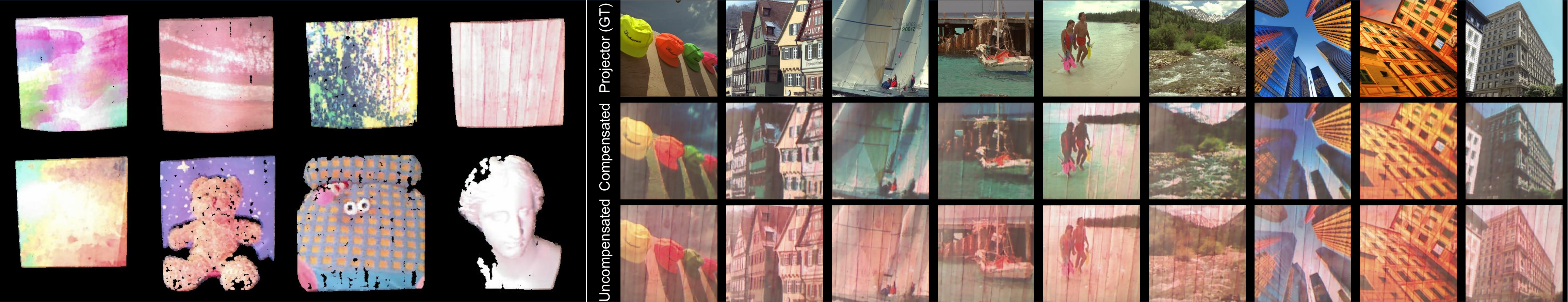}
\caption{Proof-of-concept applications of LICOA.
\textit{Left}: RGB colored point clouds reconstructed from structured light sequences for several surfaces and objects.
\textit{Right, from top to bottom}: desired projection content, compensated camera observations, and uncompensated camera observations on one textured surface.
The compensation images are rectified to projector coordinates for visual comparison.
}
  \label{fig:applications}
\end{figure*}

As shown in~\cref{tab:ablation}, the full configuration achieves the highest SR and SPL, at $60.00\%$ and $37.09\%$, respectively. Removing the external camera reduces SR to $26.67\%$ and SPL to $13.47\%$, while removing the onboard camera reduces them to $30.00\%$ and $24.61\%$. Thus, both the global scene context and robot-centric visual feedback make substantial contributions. Although the $6.57\,\mathrm{s}$ MT without the external camera is lower than the full configuration's $12.20\,\mathrm{s}$, it is conditioned on only eight successful rollouts and therefore does not indicate better overall performance. Removing the onboard camera instead increases MT over successful trials to $29.07\,\mathrm{s}$.
The action component ablation likewise supports coordinated control. Mirror only reaches $30.00\%$ SR, whereas Chassis only reaches $13.33\%$; both remain well below the full configuration's $60.00\%$. Their per-task SRs also differ: Mirror only obtains $0\%$, $40\%$, and $50\%$, while Chassis only obtains $30\%$, $0\%$, and $10\%$ on the three tasks. This task dependence shows that neither actuator alone provides consistent reachability and placement across target configurations. The mirror-only SPL equals its SR because successful rollouts with a stationary chassis receive unit path efficiency under the zero-motion convention; its $45.71\,\mathrm{s}$ MT nevertheless remains substantially above the full configuration's $12.20\,\mathrm{s}$. The Suppl.\ Material further studies parameter adaptation and training data scale under a fixed offline validation protocol. It shows how the matched simulation corpus supports adaptation, which cannot substitute for the real-world evaluation results above.

%% file: tables/comparison.tex

\begin{table}[t]
  \centering
  \caption{Real-world evaluation results for three pretrained action model families connected through the shared perception and optical action interface and a $\pi_{0.5}$-scratch control. SR is averaged over all $12$ tasks ($10$ rollouts per task; $120$ rollouts per action model). SPL is averaged over the same $120$ rollouts, while MT is averaged over successful rollouts only. The best and second-best results are shown in \textbf{bold} and \underline{underlined}, respectively.}
  \label{tab:main_comparison}
  \setlength{\tabcolsep}{5.6pt}
  \begin{tabular}{@{}lrrrr@{}}
    \toprule
    \textbf{Method} & \textbf{Success} & \textbf{SR ($\%$)} $\uparrow$
    & \textbf{SPL ($\%$)} $\uparrow$ & \textbf{MT ($\mathrm{s}$)} $\downarrow$ \\
    \midrule
    $\pi_{0.5}$-scratch & $10/120$ & $8.33$ & $5.15$ & \underline{$47.65$} \\
    $\pi_{0.5}$~\cite{PhysicalIntelligence2025pi05}
      & $69/120$ & $\textbf{57.50}$ & $\textbf{40.64}$ & $\textbf{43.44}$ \\
    $\pi_0$-FAST~\cite{Pertsch2025fast}
      & $21/120$ & $17.50$ & $15.81$ & $139.38$ \\
    Fast-WAM~\cite{yuan2026fastwam}
      & $35/120$ & \underline{$29.17$} & \underline{$21.76$} & $64.72$ \\
    \bottomrule
  \end{tabular}
\end{table}

%% file: tables/ablation.tex

\begin{table}[t]
  \centering
  \caption{Camera and action component ablations for $\pi_{0.5}$ on three novel real-world tasks ($10$ rollouts per task; $30$ per configuration). Camera variants are retrained without the indicated view, whereas actuation variants disable the indicated component only at deployment. All variants retain the robot control state and projection-scale command. The images below show Task~$13$, ``project white light onto the umbrella.''; Task~$14$, ``project white light onto the brown bag.''; and Task~$15$, ``project white light onto the white statue.''}
  \label{tab:ablation}
  \footnotesize
  \setlength{\tabcolsep}{2.1pt}
  \begin{tabular}{@{}lcccclrrr@{}}
    \toprule
    \textbf{Config.}
      & $\I^o$ & $\I^e$ & \textbf{Chassis} & \textbf{Mirror}
      & \textbf{Success}
      & \textbf{SR} $\uparrow$
      & \textbf{SPL} $\uparrow$
      & \textbf{MT} $\downarrow$ \\
    \midrule
    Full          & \checkmark & \checkmark & \checkmark & \checkmark & $18/30$ & \textbf{$60.00$} & \textbf{$37.09$} & $12.20$ \\
    W/o onb. cam. & $\times$   & \checkmark & \checkmark & \checkmark & $9/30$  & $30.00$ & $24.61$ & $29.07$ \\
    W/o ext. cam. & \checkmark & $\times$   & \checkmark & \checkmark & $8/30$  & $26.67$ & $13.47$ & \textbf{$6.57$} \\
    Mirror only   & \checkmark & \checkmark & $\times$   & \checkmark & $9/30$  & $30.00$ & $30.00$ & $45.71$ \\
    Chassis only  & \checkmark & \checkmark & \checkmark & $\times$   & $4/30$  & $13.33$ & $12.04$ & $23.86$ \\
    \bottomrule
  \end{tabular}

  \includegraphics[width=\linewidth]{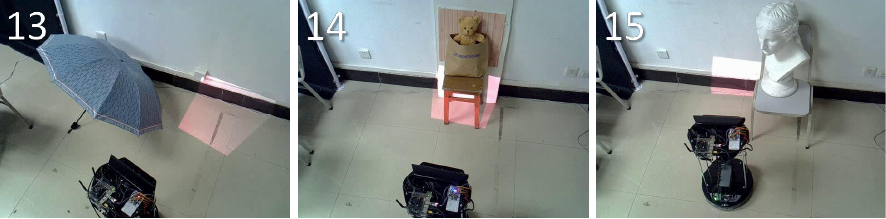}
\end{table}

%% file: sections/05_discussion.tex
\section{Discussion}
\label{sec:discussion}

The central scope of LICOA is physical placement: it interprets a spatial instruction and controls where light arrives and how large its footprint becomes. The uniform white rectangular projection used in the real-world evaluation protocol isolates this capability from content. Within that scope, the experiments support the complete causal chain: the shared interface connects pretrained action models to the platform, both camera views contribute complementary context, and the chassis and mirror provide complementary motion scales.

\subsection{Applicability to other settings}

\begin{figure}[h]
  \centering
  \includegraphics[width=\linewidth]{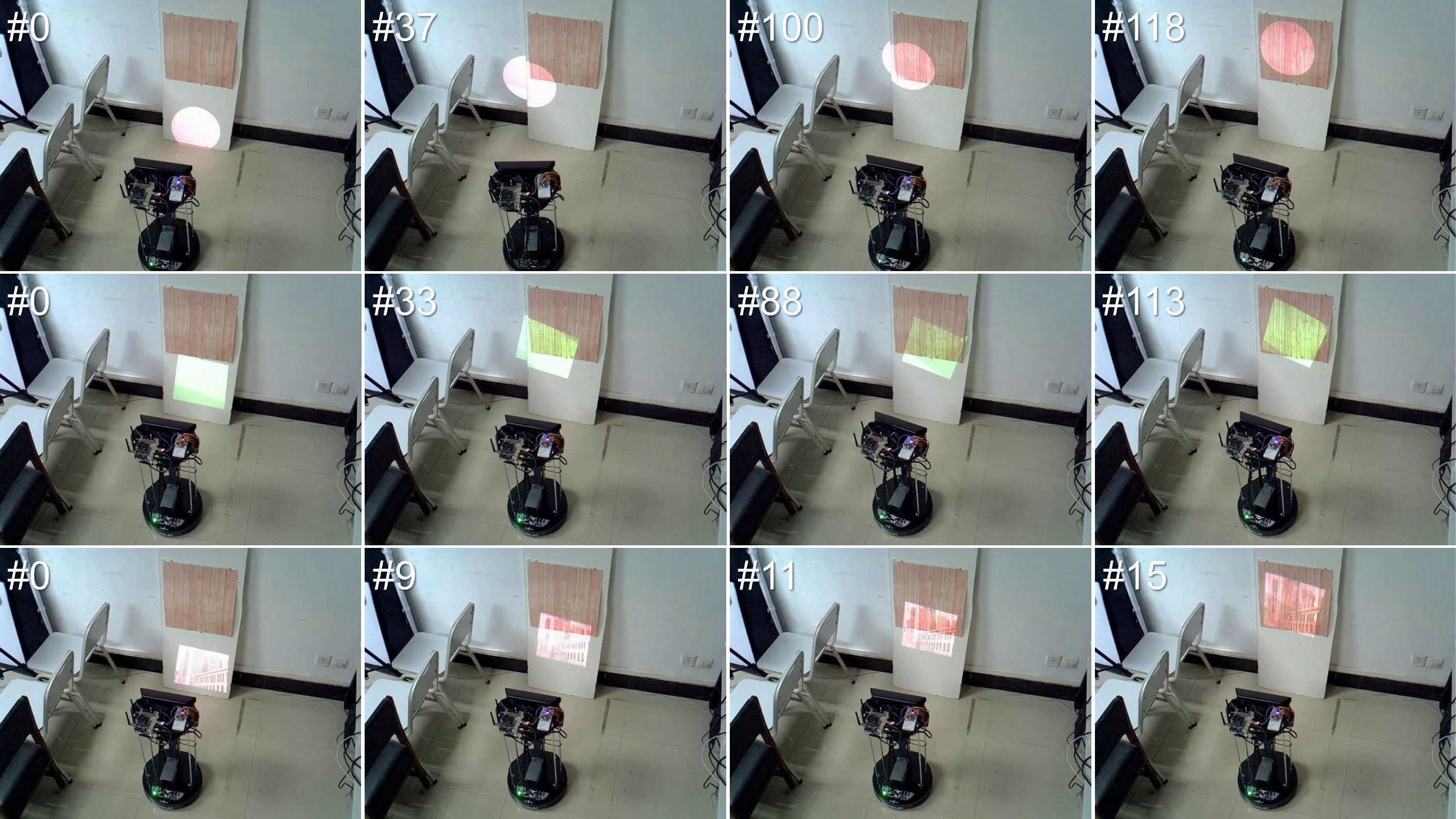}
  \caption{Qualitative rollouts of the full $\pi_{0.5}$ configuration with varied projection content. Each row shows selected frames from one rollout. From top to bottom, the instructions are ``project circle light onto the wooden surface,'' ``project green light onto the wooden surface,'' and ``project image onto the wooden surface.''}
  \label{fig:applicability}
\end{figure}

To explore learned placement beyond the projection appearance used for training, we use the same full $\pi_{0.5}$ model from the preceding experiments. The model was fine-tuned only on the matched simulation corpus with a uniform white rectangular projection. At deployment, we replace this content with a circular light, green light, or an image, without additional fine-tuning. As shown in~\cref{fig:applicability}, the model places each content on the wooden surface. These qualitative rollouts provide initial evidence that the learned placement behavior can accommodate projection appearances beyond the fine-tuning corpus, suggesting potential for simple semantic projection and motivating the downstream optical applications explored next.

\subsection{Additional applications}

Building on this potential, we test two downstream operations with the placed projector-camera pair. These proof-of-concept trials follow a \emph{place-then-operate} workflow: placement first establishes an operating pose, after which task-specific projection and processing support structured light reconstruction or projector compensation.

\noindent\textbf{Structured light 3D reconstruction.}
After LICOA places light on a target, the chassis and mirror remain fixed while the projector displays complementary Gray-code and four-step phase-shifting patterns~\cite{srinivasan1984automated,huang2021a,yamazaki2011selfcalibration}. The onboard Kinect records each pattern sequence, and the Jetson decodes projector-camera correspondences and triangulates the colored point clouds shown in~\cref{fig:applications}. These examples demonstrate local geometry acquisition after repositioning.

\noindent\textbf{Projector compensation.}
We also demonstrate compensation on one textured surface after the platform is fixed. A complementary Gray-code scan provides the geometric warp as in~\cite{Deng2025LAPIG}, and the photometric component of a pretrained CompenNeSt~\cite{huang2022CompenNeSt++} is fine-tuned on eight projected images. The scan, fine-tuning, compensated image generation, and physical reprojection run on the Jetson. This result is a single-pose proof of concept and remains separate from the real-world evaluation protocol.

\subsection{Limitations and future work}

The current formulation intentionally limits the scope of the problem. First, using a uniform white rectangular projection is sufficient for evaluating spatial placement but does not address the generation, layout, or readability of rich projected content. Second, the external camera simplifies supervision and evaluation, which makes the current setup more instrumented than a fully self-contained consumer system. Third, the projection result remains sensitive to surface material, ambient light, camera exposure, and actuator repeatability. Fourth, the proposed action space is continuous and hardware-specific, so generalization across different projectors, mirrors, mobile chassis, and room scales will require additional calibration or adaptation data. Fifth, the current procedure used to construct termination supervision does not produce fully reliable termination labels for every demonstration. Consequently, the learned termination output is not always dependable.
More reliable termination annotation and explicit supervision of task completion over time remain important to be explored.

Future work should extend on both system and learning. On the system side, richer projected content, onboard-only state estimation, a higher-speed 2DoF steerable mirror, and more accurate projection scale control can make the platform closer to practical SAR applications. On the learning side, larger VLA or WAM models can be fine-tuned with mixed real and simulated data, and self-supervised exploration can be used to learn how commands for the differential-drive chassis and 2DoF steerable mirror move the projection footprint. A longer-term goal is to develop spatial computing agents that can decide not only what information to show, but also where and how light should be physically routed in the environment.

%% file: sections/06_conclusion.tex
\section{Conclusion}
\label{sec:conclusion}

This work introduces LICOA, a mobile SAR system that treats projected light as an instruction-conditioned optical action. By coupling room-scale chassis motion with rapid 2DoF mirror steering and projection-scale control, LICOA decides how to move and where to place light without fixed projector poses, predefined targets, or scripted trajectories. The shared perception and optical action interface connects diverse pretrained action models to the same platform. Trained only on a matched simulation corpus of $150$ spatial placement tasks and $7{,}500$ episodes, adapted $\pi_{0.5}$ achieves $57.50\%$ SR and $40.64\%$ SPL across $480$ real-world rollouts, versus $8.33\%$ and $5.15\%$ from scratch. Ablations further validate the complementary sensing and actuation design. Together, these results turn projection from a configured display into an embodied capability, opening a path toward autonomous and adaptive SAR.

%% file: supp/supp.tex
\newpage

\clearpage
\onecolumn

\begin{center}
{\sffamily\LARGE\bfseries
LICOA: Learning Instruction-Conditioned Optical Actions for Mobile Spatial Augmented Reality \\
\vspace{0.3em}
\normalsize\textit{-- Supplementary Materials --}
}
\end{center}

\renewcommand{\theequation}{S\arabic{equation}}
\renewcommand{\thefigure}{S\arabic{figure}}
\renewcommand{\thetable}{S\arabic{table}}
\setcounter{equation}{0}
\setcounter{figure}{0}
\setcounter{table}{0}
\setcounter{section}{0}

\input{supp/sections/01_hardware_control}
\input{supp/sections/02_simulation_dataset}
\input{supp/sections/03_experiments}

%% file: supp/sections/01_hardware_control.tex

\section{Physical Platform and Embedded Control}
\label{sec:supp_gimbal_control}

The physical implementation follows the design with two motion scales in the main paper. The differential-drive chassis carries the projector across the environment, while the steerable mirror with two degrees of freedom (2DoF) redirects the beam locally and projection scale changes its footprint. The embedded system must therefore deliver chassis, mirror, and scale commands through separate hardware paths while preserving the shared action components exposed to the action model.

\subsection{System interfaces and power}

The NVIDIA Jetson Nano is the robot-mounted embedded computer. It acquires the onboard RGB stream from the Microsoft Kinect v1 over USB, communicates with the Kobuki differential-drive chassis over a dedicated USB serial connection, and exchanges pan and tilt targets for the 2DoF steerable mirror and telemetry with the STM32 through a separate universal asynchronous receiver-transmitter (UART) interface over USB. The Jetson sends the robot stream, comprising the latest onboard RGB image, its timestamp, and the robot control state, to the host PC. The USB-connected Logitech C270 supplies the external camera stream. The host PC synchronizes the two streams, sends the resulting observation as an inference request to the server, receives a $16$-step action chunk, and relays each executed six-dimensional per-step action to the Jetson. The Jetson then dispatches commands to the Kobuki, STM32, and custom laser projector. \Cref{tab:supp_power} summarizes the power distribution.

\subsection{Custom laser projector}

The custom laser projector receives HDMI output from the Jetson. A Lontium LT8619B bridge converts HDMI to $24$-bit RGB888 for the Sony CXN0102-based projection electronics, which coordinate Sony image processing and laser control with the RGB laser sources and a scanner based on microelectromechanical systems (MEMS) to produce the projected image. Projection scale is changed by rendering the corresponding pattern over the same HDMI path.

\subsection{2DoF steerable mirror wiring}

The two physical USB connections between the Jetson and the STM32 assembly serve different purposes. The CH340 USB-to-UART adapter carries bidirectional commands and telemetry, whereas the direct USB connection supplies STM32 logic power. Motor power is independent of both links and is delivered from the $12\,\mathrm{V}$ rail to the two SimpleFOC Mini driver boards.

\subsection{Jetson Nano command bridge}

The NVIDIA Jetson Nano is the robot-mounted bridge between the remote host PC control client and the action components. Its bridge service receives each six-dimensional per-step action as a ZeroMQ (ZMQ) message containing the linear and angular velocities $(v,\omega)$ for the differential-drive chassis, pan and tilt targets $(q_{\mathrm{p}},q_{\mathrm{t}})$ for the 2DoF steerable mirror, projection scale $\rho$, and termination indicator $b$. It clamps the two mirror targets to the platform limits, $q_{\mathrm{p}}\in[-\pi/2,\pi/2]$ and $q_{\mathrm{t}}\in[0,\pi/2]$, dispatches chassis, mirror, and projection-scale commands through their respective interfaces, and zeros the chassis velocities when $b>0.5$. The pan-tilt serial controller converts the two targets from semantic joint coordinates in radians to the absolute angles expected by the firmware. Let $\phi_{\mathrm{X}}^0$ and $\phi_{\mathrm{Q}}^0$ be the configurable zero offsets; denote the resulting firmware angles by $\phi_{\mathrm{X}}$ and $\phi_{\mathrm{Q}}$; and let $\operatorname{wrap}_{360}$ map an angle in degrees to $[0,360)$. The pan-axis mapping is
\begin{equation}
  \phi_{\mathrm{X}}=\operatorname{wrap}_{360}\!\left(
    \phi_{\mathrm{X}}^0-\frac{180}{\pi}q_{\mathrm{p}}\right).
  \label{eq:supp_pan_mapping}
\end{equation}
The tilt-axis mapping is
\begin{equation}
  \phi_{\mathrm{Q}}=\operatorname{wrap}_{360}\!\left(
    \phi_{\mathrm{Q}}^0+\frac{180}{\pi}q_{\mathrm{t}}\right).
  \label{eq:supp_tilt_mapping}
\end{equation}
The opposite sign in~\cref{eq:supp_pan_mapping}, compared with~\cref{eq:supp_tilt_mapping}, accounts for the installed motor orientation.

The Jetson encodes pan (X) and tilt (Q) targets as newline-terminated ASCII commands, for example \texttt{X45.00} and \texttt{Q30.00}. A USB-to-UART adapter connects to STM32 USART1 (PA9/PA10 and common ground) at $115{,}200\,\mathrm{baud}$, 8N1; a second USB connection from the Jetson supplies STM32 logic power. Because the firmware uses a single line receive buffer, a background sender alternates the Q and X lines with a $10\,\mathrm{ms}$ offset; each axis is consequently refreshed at $50\,\mathrm{Hz}$. Newly received per-step commands overwrite the previous targets, so command latency does not build an obsolete FIFO queue. The serial interface reconnects lazily after disconnection, and a missing pan-tilt link does not block control of the differential-drive chassis. Separately, a $0.3\,\mathrm{s}$ command watchdog zeros the velocity of the differential-drive chassis; it does not reset the pan-tilt targets. The deployed path does not use a controller area network (CAN): communication is UART-over-USB from Jetson to STM32, software inter-integrated circuit (I2C) communication from STM32 to the encoders, and pulse-width modulation (PWM) plus enable signals from STM32 to the motor drivers.

\input{supp/tables/power}

\begin{figure*}[ht]
  \centering
  \includegraphics[width=0.92\textwidth]{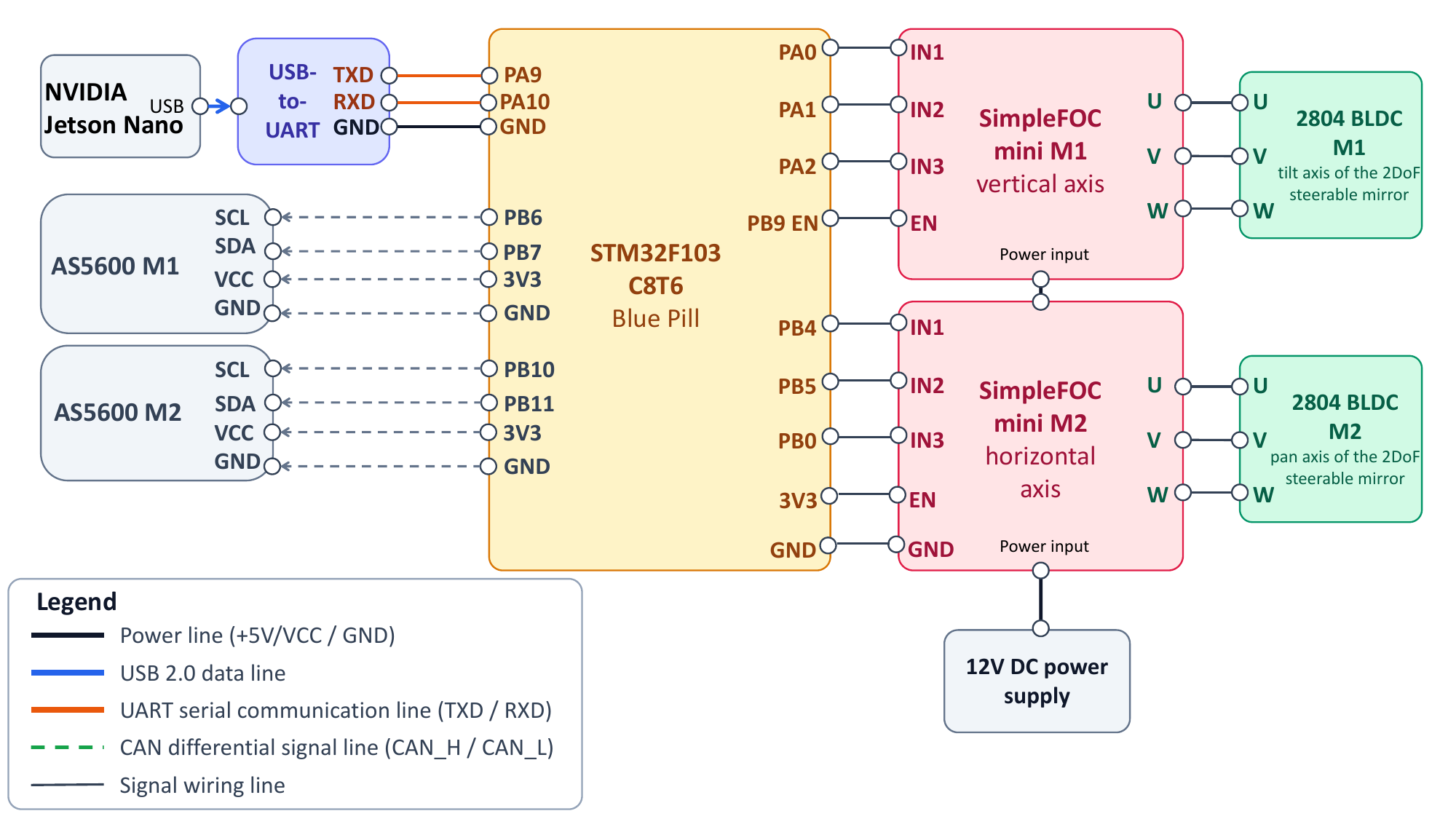}
  \caption{Electrical and control connections for the 2DoF steerable mirror and its pan-tilt servo assembly. The NVIDIA Jetson Nano mounted on the differential-drive chassis provides the USB-to-UART command and telemetry path, while a separate Jetson USB connection (not shown) supplies logic power to the STM32. The STM32 closes both motor loops locally using AS5600 encoder feedback and drives the two brushless direct-current (BLDC) driver boards with three-phase pulse-width modulation (PWM). The $12\,\mathrm{V}/1\,\mathrm{A}$ rail from battery pack~$2$ supplies the two SimpleFOC Mini drivers and pan-tilt motors.}
  \label{fig:supp_hardware_connection}
\end{figure*}

\subsection{STM32 motor control}

The USB link carries targets and diagnostic telemetry, not real-time servo waveforms. An STM32F103C8T6 parses the X/Q commands and closes the two pan-tilt motor loops locally at approximately $260\,\mathrm{Hz}$. Each $2804$ brushless direct-current (BLDC) axis uses a $12$-bit AS5600 magnetic encoder connected by software I2C (PB6/PB7 for the vertical axis and PB10/PB11 for the horizontal axis). Startup calibration estimates sensor direction and electrical zero; during operation, a five-sample median filter suppresses angular outliers.

For streamed commands, the firmware interpolates each internal target at $0.30\,\mathrm{rad/s}$ using the measured control loop interval, so its motion profile remains stable when the loop rate varies. Both axes then use continuous position/velocity feedback to compute the q-axis voltage command. The deployed vertical-axis path does not use the legacy staged open-loop fast approach mode. The requested voltage is converted to space-vector pulse-width modulation (SVPWM) at approximately $25\,\mathrm{kHz}$ and sent as three PWM phases plus an enable signal to each SimpleFOC Mini driver. Both drivers use a $12\,\mathrm{V}$ supply, with firmware voltage limits of $6.5\,\mathrm{V}$ for the vertical motor and $6.0\,\mathrm{V}$ for the horizontal motor. Encoder angle, estimated velocity, loop rate, and error flags are returned over USB for diagnostics. Thus, network and Jetson timing determine when a new setpoint arrives, whereas the STM32 maintains the high-rate electromechanical feedback loop shown in
\cref{fig:supp_hardware_connection}.

%% file: supp/tables/power.tex

\begin{table*}[ht]
  \centering
  \caption{Power distribution of the physical LICOA platform. Capacities and output ratings are the nameplate specifications of the deployed supplies.}
  \label{tab:supp_power}
  \small
  \renewcommand{\arraystretch}{1.12}
  \begin{tabularx}{\textwidth}{@{}
    >{\raggedright\arraybackslash}p{0.22\textwidth}
    >{\raggedright\arraybackslash}p{0.14\textwidth}
    >{\raggedright\arraybackslash}p{0.25\textwidth}
    >{\raggedright\arraybackslash}X@{}}
    \toprule
    \textbf{Source} & \textbf{Output} & \textbf{Load} & \textbf{Purpose} \\
    \midrule
    Battery pack~$1$ ($6{,}000\,\mathrm{mAh}$)
      & Chassis supply
      & Kobuki mobile chassis
      & Differential-drive locomotion \\
    Battery pack~$1$ ($6{,}000\,\mathrm{mAh}$)
      & $12\,\mathrm{V}/1.5\,\mathrm{A}$
      & Microsoft Kinect v1
      & Onboard RGB camera power; image data use USB \\
    Battery pack~$2$ ($30{,}000\,\mathrm{mAh}$)
      & $5\,\mathrm{V}/4\,\mathrm{A}$
      & NVIDIA Jetson Nano
      & Embedded computing \\
    Battery pack~$2$ ($30{,}000\,\mathrm{mAh}$)
      & $12\,\mathrm{V}/1\,\mathrm{A}$
      & Two SimpleFOC Mini drivers
      & Actuation of the 2DoF steerable mirror \\
    Battery pack~$2$ ($30{,}000\,\mathrm{mAh}$)
      & $5\,\mathrm{V}/2\,\mathrm{A}$
      & Custom laser projector
      & Projection \\
    NVIDIA Jetson Nano
      & USB bus power
      & STM32F103C8T6
      & Controller logic power, separate from USB-to-UART data \\
    Host PC
      & USB bus power
      & Logitech C270
      & External RGB camera data and power \\
    \bottomrule
  \end{tabularx}
\end{table*}

%% file: supp/sections/02_simulation_dataset.tex

\section{Matched Simulation Corpus Construction}
\label{sec:supp_simulation_dataset}

The main paper identifies the absence of training data and common tasks for instruction-conditioned optical action as a central challenge. We address it by reproducing the physical LICOA embodiment, camera roles, robot control state, and action components in simulation, then collecting demonstrations under the same interface used for real-world deployment.

We construct the simulated environment using NVIDIA Isaac Sim 6.0.1. Indoor geometry and materials are loaded from the first $10$ InteriorAgent universal scene description (USD) scenes at a scale of one meter per stage unit. Each InteriorAgent scene is a complete, large indoor environment. We define five fixed external camera viewpoints in each scene and construct three spatial placement tasks for each external camera viewpoint, yielding $150$ tasks in total. For each task, we add the LICOA USD articulation, a PhysX scene, and the task-specific initial robot pose. Environment rigid bodies are frozen during collection, and static triangle-mesh collision proxies preserve the scene geometry used for navigation and optical interaction. The onboard RGB camera remains attached to the robot, whereas the external RGB camera remains at the corresponding fixed pose in the world frame. The principal recording settings are summarized in
\cref{tab:supp_isaac_config}.

\begin{table}[ht]
  \centering
  \caption{Isaac Sim configuration used to match the LICOA physical embodiment and construct its matched simulation corpus.}
  \label{tab:supp_isaac_config}
  \footnotesize
  \renewcommand{\arraystretch}{1.12}
  \begin{tabularx}{\columnwidth}{@{}
    >{\raggedright\arraybackslash}p{0.28\columnwidth}
    >{\raggedright\arraybackslash}X@{}}
    \toprule
    \textbf{Setting} & \textbf{Configuration} \\
    \midrule
    Simulator
      & NVIDIA Isaac Sim 6.0.1; PhysX fixed-step simulation \\
    Scene assets
      & First $10$ InteriorAgent USD indoor scenes; each a complete, large indoor
        environment; $1$ stage unit $=1\,\mathrm{m}$; frozen
        environment rigid bodies with static triangle-mesh collision proxies \\
    Differential-drive chassis
      & Differential drive; wheel radius $0.035\,\mathrm{m}$; wheelbase $0.230\,\mathrm{m}$;
        $\lvert v\rvert\leq0.7\,\mathrm{m/s}$ and
        $\lvert\omega\rvert\leq3.14\,\mathrm{rad/s}$ \\
    2DoF steerable mirror and projection scale
      & Pan range $[-\pi/2,\pi/2]$; tilt range $[0,\pi/2]$; projection scale
        $[0,100]\%$ \\
    Virtual projector
      & Textured square RectLight; uniform white $256\!\times\!256$ source image; intensity
        $80$; throw-ratio parameter $6.0$ \\
    External and onboard RGB images
      & Onboard RGB camera and five fixed external camera viewpoints per
        scene; both streams rendered at $640\!\times\!480$ \\
    Sampling
      & Fixed $15\,\mathrm{Hz}$ application, timeline, and PhysX steps; synchronized
        RGB image, robot control state, and action records \\
    Offline rendering
      & RTX Real-Time renderer; deep learning anti-aliasing (DLAA); four render subframes per recorded state;
        motion blur disabled; $640\!\times\!480$ output \\
    Appearance variants
      & One original rendering and four randomized renderings per source
        trajectory \\
    \bottomrule
  \end{tabularx}
\end{table}

We collect source trajectories by teleoperating the differential-drive chassis, the steerable mirror with two degrees of freedom (2DoF), and the projection scale. Only demonstrations that the operator accepts as successfully satisfying the instruction are committed as source trajectories, while failed and timed-out attempts are discarded. Isaac Sim uses fixed time stepping at $15\,\mathrm{Hz}$, so every saved observation is aligned with one complete physics step. We first save the full simulator state together with the synchronized trace of robot control states and actions and then replay it with physics paused to render the external and onboard RGB images at the recorded poses. This separates interactive demonstration collection from offline image generation while preserving the exact source action sequence. At each recorded simulator state, four rendering subframes are used: two interpolate motion from the preceding state and two render the final fixed pose. Only the image at the recorded pose
is stored.

For every source trajectory, we retain the original appearance and generate four deterministic appearance variants. For each variant, the intensities of environment lights other than the projector are multiplied by a factor sampled uniformly from $[0.65,1.35]$, and their color temperature is sampled uniformly from $[3500,7000]\,\mathrm{K}$. Floor materials are sampled from eight official Isaac Sim wood, stone, tile, and carpet materials, while wall materials are sampled from four official paint, ceramic, porcelain, and brick materials. Randomization is authored only in a disposable replay layer; the projection light, scene geometry, camera poses, and source action sequence remain unchanged. The random seed and selected lighting and material settings are stored with each rendered episode.

%% file: supp/sections/03_experiments.tex

\section{Action Model Adaptation and Training Details}
\label{sec:supp_main_training}

The shared perception and optical action interface keeps the physical inputs and outputs fixed while allowing each action model to retain its own representation and objective. This section reports the model-specific conversions and optimization settings used to test that interface; none of the runs uses real-world LICOA training data.

\input{supp/tables/main_training_recipes}

The settings in~\cref{tab:supp_main_training_recipes} are the configurations stored with the checkpoints used in the main comparison. Each listed run uses one GPU. All four configurations train for $30{,}000$ updates ($960{,}000$ sample draws) with batch size $32$ and seed $0$. Online image transformations are disabled. The external and onboard RGB images are renamed to the feature keys required by each action model, and any unused third visual slot is supplied as a masked blank input.

The $\pi_{0.5}$ run performs full fine-tuning: its PaliGemma backbone, Gemma-300M action expert, and action/time projections are all trainable. The $\pi_0$-FAST run also performs full fine-tuning and uses neither low-rank adaptation (LoRA) nor another parameter-efficient fine-tuning (PEFT) module. It tokenizes the physical action chunk with the FAST tokenizer and optimizes causal cross-entropy. Fast-WAM consumes the five-dimensional robot control state and six-dimensional action directly. Its external and onboard RGB images are concatenated at $224\times448$ resolution, its video sequence contains $17$ frames, and its video and action losses have equal weight. The $\pi_{0.5}$-scratch control uses the same optimizer and architecture as $\pi_{0.5}$ but does not load a pretrained checkpoint.

For the camera input ablation in the main paper, both variants reuse the full $\pi_{0.5}$ recipe above. Dataset preparation removes either the external or onboard RGB image, and the corresponding input key mapping for the action model is omitted; the other RGB image, robot control state, action interface, update budget, and optimizer remain unchanged.

\section{Offline Analysis of Adaptation and Data Scale}
\label{sec:supp_training_ablations}

These analyses characterize two supporting questions behind the real-world evaluation protocol: how much of the pretrained action model must be adapted, and how the amount and diversity of simulation data affect action prediction. We conduct two offline ablations with $\pi_{0.5}$: one varies the set of trainable parameters, and the other varies training data scale along two independent axes. These experiments do not use real-world rollouts. Their metric is the flow-matching loss on one fixed held-out set, which makes the training configurations inexpensive to compare but does not measure closed-loop recovery or task success. We therefore treat these results as optimization and data-efficiency evidence rather than a substitute for the success rate (SR), success weighted by path length (SPL), and mean completion time (MT) results in the main paper.

\subsection{Corpus and offline validation split}
\label{sec:supp_training_split}

The matched simulation corpus uses the first $10$ InteriorAgent scenes, each of which is a complete, large indoor environment. In each scene, we define five fixed external camera viewpoints and construct three tasks per viewpoint. The resulting $10\times5\times3=150$ \emph{spatial placement tasks} each have $10$ manually collected source trajectories. We render every trajectory with five appearance settings: the original appearance and four variants with randomized lighting intensity, lighting color temperature, wall materials, and floor materials. This gives $50$ episodes per task and $7{,}500$ episodes ($919{,}950$ frames at $15\,\mathrm{Hz}$) in total. The $1{,}465$ task strings reported in the LeRobot metadata are language instruction instances and should not be interpreted as distinct spatial placement tasks.

This holdout is used only to compare offline validation loss; it is not a benchmark for generalization across scenes or instructions. All $10$ InteriorAgent scenes contribute data to the training split used by the main experiments and ablations, and the configurations using the complete corpus include all $150$ spatial placement tasks. We split by source trajectory rather than by rendered episode. With \texttt{split\_seed=0}, one of the $10$ source trajectories from every spatial placement task and all five of its appearance variants are assigned to validation; the other nine groups are assigned to training. Hence, the fixed validation set covers all $150$ tasks and contains $750$ episodes and $96{,}510$ frames. The full training split contains $6{,}750$ episodes and $823{,}440$ frames. Grouping the appearance variants prevents the same source action sequence from appearing on both sides of the split. Every configuration below is evaluated on this identical validation set. Real-world generalization is evaluated separately in previously unseen real-world environments with new instructions.

The action model receives timestamp-synchronized onboard and external RGB images; the stored projector input stream is excluded. In the $\pi_{0.5}$ input convention, these are mapped to \texttt{base\_0\_rgb} and \texttt{left\_wrist\_0\_rgb}, respectively, while the unused right-wrist slot is supplied as a masked blank image. The five-dimensional robot control state is $(v,\omega,q_{\mathrm{p}},q_{\mathrm{t}},\rho)$, where $q_{\mathrm{p}}$ and $q_{\mathrm{t}}$ are the pan and tilt targets and $\rho$ is the projection scale rather than an optical field-of-view angle. The six-dimensional action contains the corresponding commanded values and the additional termination indicator. The first and ninety-ninth percentiles computed over the complete $7{,}500$ episode corpus are used to map the robot control state and action channels to $[-1,1]$; these same statistics are deliberately held fixed in every data scale run. Appearance variation is generated offline, and no additional online image augmentation is applied.

\subsection{Training and validation protocol}
\label{sec:supp_training_protocol}

\input{supp/tables/training_protocol}

As summarized in~\cref{tab:supp_training_protocol}, all runs use LeRobot's implementation of $\pi_{0.5}$. Its vision-language backbone is the PaliGemma/Gemma-2B variant, and its action expert is the Gemma-300M variant. The logged action model contains $4{,}143{,}404{,}816$ parameters in total. Its adapter pads both the five-dimensional robot control state and the six-dimensional action with zeros to form LeRobot's $32$-dimensional tensors. The loss is computed only over the six physical LICOA action channels, and only those channels are kept after inference. The $\pi_0$-FAST adapter instead serializes the normalized robot control state through its prompt preprocessor and passes the unpadded six-dimensional continuous action chunk to the FAST tokenizer. Fast-WAM consumes the five-dimensional robot control state and six-dimensional action directly; its embodiment-dependent state encoder and action input/output layers are resized accordingly, while all shape-compatible pretrained weights are loaded. The action horizon is $H=16$, corresponding to approximately $1.067\,\mathrm{s}$ at $15\,\mathrm{Hz}$. No additional temporal validity mask is introduced.

Using the main paper notation with $m=\pi_{0.5}$, let $B$ be the validation batch size and index its samples by $n\in\{1,\ldots,B\}$. Sample $n$ contains target action chunk $\mathbf{A}_{t_n}^*$ and conditioning input $x_{t_n}^{(m)}$, where $t_n$ is its control time index. Here, $H=16$ is the action horizon and $d_a=6$ is the physical action dimension, so $\mathbf{A}_{t_n}^*\in\mathbb{R}^{H\times d_a}$. For each sample, draw Gaussian noise $\boldsymbol{\epsilon}_n$ of the same shape, with independent standard normal entries, and a flow time variable $\lambda_n\sim\operatorname{Beta}(1.5,1)$ (subsequently mapped to $0.999\lambda_n+0.001$). Following the implementation's flow time convention, the noisy action chunk is
\begin{equation}
  \widetilde{\mathbf{A}}_{t_n,\lambda_n}
  =\lambda_n\boldsymbol{\epsilon}_n
  +(1-\lambda_n)\mathbf{A}_{t_n}^*.
  \label{eq:supp_noisy_action_chunk}
\end{equation}

The corresponding flow target is
\begin{equation}
  \mathbf{r}_{t_n,\lambda_n}
  =\boldsymbol{\epsilon}_n-\mathbf{A}_{t_n}^*.
  \label{eq:supp_flow_target}
\end{equation}

Let $f_{\theta_m}^{(m)}$ denote the vector-field predictor native to model $m$, with $r\in\{1,\ldots,H\}$ indexing tensor positions along the action horizon and $k$ indexing action channels. We report its mean-squared flow-matching objective
\begin{equation}
  \begin{aligned}
    \mathcal{L}_{\mathrm{valid}}
    &= \frac{1}{BHd_a}
    \sum_{n=1}^{B}\sum_{r=1}^{H}\sum_{k=1}^{d_a} \\
    &\quad \left(
      f_{\theta_m}^{(m)}\!\left(\widetilde{\mathbf{A}}_{t_n,\lambda_n},
      x_{t_n}^{(m)},\lambda_n\right)_{rk}
      -\mathbf{r}_{t_n,\lambda_n,rk}
    \right)^2.
  \end{aligned}
  \label{eq:supp_validation_loss}
\end{equation}
The construction in~\cref{eq:supp_noisy_action_chunk,eq:supp_flow_target} defines the target used by the validation loss in~\cref{eq:supp_validation_loss}. The entire $96{,}510$-frame validation set is evaluated every $5{,}000$ updates. A fixed noise/time random number stream is reused for corresponding validation batches across configurations to reduce Monte-Carlo variation. We use the loss at the common $30{,}000$-update endpoint as the primary table entry and retain all six checkpoints for convergence plots. Runs share one seed; this controls the comparison but does not constitute a multi-seed uncertainty estimate.

\subsection{Parameter adaptation scope}
\label{sec:supp_adaptation_range}

\Cref{tab:supp_parameter_adaptation} compares three configurations. In \emph{Projection heads}, only the action input/output projections and time multilayer perceptron (MLP) are trainable. \emph{Action expert}, a parameter-efficient alternative, also updates the Gemma action expert while freezing the vision-language backbone. \emph{Full} updates the complete pretrained action model and is the recipe used by $\pi_{0.5}$ in the main real-world comparison. The logged parameterization comprises $3{,}449{,}982{,}704$ parameters in the vision-language and image backbones, $691{,}256{,}320$ action expert parameters, and $2{,}165{,}792$ action/time projection parameters. In the Action expert configuration, the first group is frozen and the latter two are optimized.

\input{supp/tables/parameter_adaptation}

Alongside endpoint validation loss, we record peak allocated GPU memory and wall-clock runtime. Peak memory is measured with \texttt{torch.cuda.max\_memory\_allocated}, not reserved memory from \texttt{nvidia-smi}, and we take the maximum value logged over all training updates. Runtime is the W\&B \texttt{\_runtime} from run initialization to completion, so it includes action model and dataset construction in addition to optimization, validation, and checkpointing.

Updating the action expert reduces the endpoint loss by $22.3\%$ relative to training only the projection heads ($0.3030$ versus $0.3901$). Full fine-tuning improves the endpoint by a further $0.0026$, while increasing the trainable fraction from $16.74\%$ to $100\%$ and peak allocated memory from $21.79$ to $40.23\,\mathrm{GiB}$. The Projection heads and Action expert runs both take about $15.3\,\mathrm{h}$, whereas Full takes $19.6\,\mathrm{h}$. Thus, Action expert is a parameter-efficient alternative that nearly matches Full on this offline metric with $83.3\%$ fewer trainable parameters and $45.8\%$ lower peak memory; all main paper $\pi_{0.5}$ results nevertheless use Full fine-tuning.

\subsection{Data scale analysis}
\label{sec:supp_data_scale}

We vary data scale in two complementary ways using \emph{Full} fine-tuning in all completed data scale runs. The \emph{trajectory density} axis keeps all $150$ tasks and uses $10$, $20$, $30$, $40$, or $50$ available episodes per task. Because five episodes per task belong to the fixed validation group, these levels contain $1$, $3$, $5$, $7$, or $9$ training source trajectories ($5$, $15$, $25$, $35$, or $45$ rendered training episodes) per task. The subsets are nested.

The \emph{task diversity} axis fixes the maximum $50$ available episodes per task and trains on nested subsets of $10$, $50$, or $150$ spatial placement tasks. Validation always covers all $150$ tasks. Consequently, validation for the $10$- and $50$-task action models mixes seen and unseen spatial placement tasks and should not be described as a pure within-task reconstruction test.

\input{supp/tables/data_scale}

\begin{figure*}[t]
  \centering
  \includegraphics[width=\textwidth]{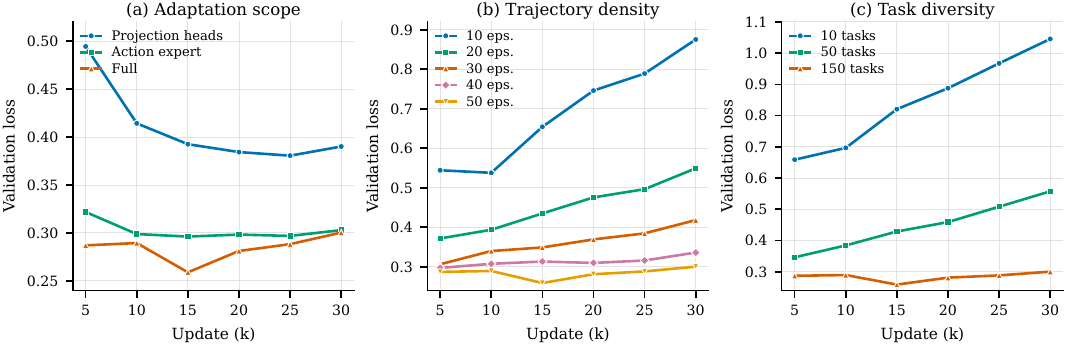}
  \caption{Offline validation loss at the six common checkpoints from $5{,}000$ to $30{,}000$ updates. The adaptation scope panel uses the complete training split. The data scale panels use full fine-tuning and vary either the available episodes per task or the number of training tasks. All curves use the same fixed validation set; lower is better. Curves combine distinct line styles and markers so that they remain identifiable without color.}
  \label{fig:supp_offline_ablation_curves}
\end{figure*}

At the common $30{,}000$-update endpoint, increasing available episodes per task from $10$ to $50$ reduces validation loss monotonically from $0.8755$ to $0.3004$, a $65.7\%$ reduction. Increasing the training task set from $10$ to $150$ likewise reduces the endpoint loss from $1.0454$ to $0.3004$, a $71.3\%$ reduction. Thus both additional trajectories for covered tasks and broader spatial task coverage substantially improve held-out action prediction under the fixed compute budget.

The checkpoint curves in \cref{fig:supp_offline_ablation_curves} also expose an important endpoint effect. The run using the complete training split reaches its lowest measured loss of $0.2589$ at $15{,}000$ updates before ending at $0.3004$. Smaller training subsets usually turn upward earlier: for example, the $10$-episode trajectory run moves from its minimum of $0.5379$ at $10{,}000$ updates to $0.8755$, and the $10$-task run moves from $0.6588$ at $5{,}000$ updates to $1.0454$. This behavior is consistent with overfitting as smaller training subsets are revisited more often. We nevertheless use the prespecified $30{,}000$-update endpoint for the table so that no configuration benefits from post-hoc checkpoint selection.

All data scale action models receive the same $30{,}000$ optimizer updates, so the study holds compute steps rather than epochs fixed. Smaller subsets are therefore revisited more often, as quantified by the effective epoch column in \cref{tab:supp_data_scale}. This design isolates whether additional trajectory density or spatial task diversity improves held-out action prediction under a common optimization budget, but it also means that data quantity and sample reuse cannot be interpreted independently. Together with the single seed and normalization statistics computed over the complete corpus, this is a stated limitation of the offline study.

%% file: supp/tables/main_training_recipes.tex

\begin{table*}[t]
  \centering
  \caption{Training recipes used for the main real-world comparison. All runs use one NVIDIA H200 GPU, $30{,}000$ optimizer updates, batch size $32$, seed $0$, BF16 computation, gradient checkpointing, and a $16$-step action horizon. ``Clip'' denotes the global gradient norm threshold. $\mathrm{lr}$ and $\mathrm{wd}$ denote learning rate and weight decay, respectively.}
  \label{tab:supp_main_training_recipes}
  \footnotesize
  \setlength{\tabcolsep}{3.5pt}
  \resizebox{\textwidth}{!}{%
    \begin{tabular}{@{}lllllll@{}}
      \toprule
      \textbf{Configuration} & \textbf{Initialization} & \textbf{Trainable scope}
        & \textbf{Native objective} & \textbf{AdamW $(\mathrm{lr},\boldsymbol{\beta},\mathrm{wd})$}
        & \textbf{Clip} & \textbf{Schedule} \\
      \midrule
      $\pi_{0.5}$ & Pretrained & Full action model & Flow matching
        & $(5\!\times\!10^{-5},(0.9,0.95),10^{-10})$ & $1$
        & $10{,}000$-step warmup, then constant \\
      $\pi_0$-FAST & Pretrained & Full action model & FAST token causal cross-entropy (CE)
        & $(2.5\!\times\!10^{-5},(0.9,0.95),10^{-10})$ & $1$
        & $1{,}000$-step warmup, cosine to $2.5\!\times\!10^{-6}$ \\
      Fast-WAM & Pretrained & Video and action experts
        & $\mathcal{L}_{\mathrm{video}}+\mathcal{L}_{\mathrm{action}}$
        & $(10^{-4},(0.9,0.999),10^{-2})$ & $10$
        & Zero-warmup cosine \\
      $\pi_{0.5}$-scratch & Random & Full action model & Flow matching
        & $(5\!\times\!10^{-5},(0.9,0.95),10^{-10})$ & $1$
        & $10{,}000$-step warmup, then constant \\
      \bottomrule
    \end{tabular}%
  }
\end{table*}

%% file: supp/tables/training_protocol.tex

\begin{table}[t]
  \centering
  \caption{Common offline training protocol.}
  \label{tab:supp_training_protocol}
  \small
  \begin{tabular}{@{}lp{0.61\columnwidth}@{}}
    \toprule
    \textbf{Setting} & \textbf{Value} \\
    \midrule
    Compute & Single NVIDIA H200 GPU per run ($143{,}771\,\mathrm{MiB}$); batch size $32$ \\
    Precision & BF16 mixed precision; gradient checkpointing; no compilation \\
    Optimizer & AdamW, $\boldsymbol{\beta}=(0.9,0.95)$, $\epsilon=10^{-8}$,
      weight decay $10^{-10}$ \\
    Learning rate & $5\!\times\!10^{-5}$; $10{,}000$-step linear warmup, then constant \\
    Update budget & $30{,}000$ steps ($960{,}000$ sample draws); gradient norm clipped to $1.0$ \\
    Logging & Log every $100$ steps; validate and save every $5{,}000$ steps \\
    Validation & Full $96{,}510$ frames; batch size $32$; no rollout evaluation \\
    Data loading & $4$ workers; prefetch factor $4$; persistent workers \\
    Software & LeRobot 0.6.0, PyTorch 2.11.0+cu128, Transformers 5.5.4,
      Accelerate 1.14.0, Datasets 4.8.5 \\
    \bottomrule
  \end{tabular}
\end{table}

%% file: supp/tables/parameter_adaptation.tex

\begin{table*}[t]
  \centering
  \caption{Parameter adaptation ablation for $\pi_{0.5}$. All runs use the full training split, the same $30{,}000$-update budget, and seed. Validation loss is measured at the common final update. Peak memory is the maximum logged CUDA tensor allocation, and runtime is the W\&B wall-clock runtime.}
  \label{tab:supp_parameter_adaptation}
  \small
  \begin{tabular}{@{}llrrrrr@{}}
    \toprule
    \textbf{Configuration}
      & \textbf{Initialization}
      & \textbf{Trainable params.}
      & \textbf{Ratio ($\%$)}
      & \textbf{Valid loss} $\downarrow$
      & \textbf{Peak memory ($\mathrm{GiB}$)} $\downarrow$
      & \textbf{Runtime ($\mathrm{h}$)} $\downarrow$ \\
    \midrule
    Projection heads & Pretrained & $2{,}165{,}792$ & $0.0523$ & $0.3901$ & $19.65$ & $15.27$ \\
    Action expert & Pretrained & $693{,}422{,}112$ & $16.7356$ & $0.3030$ & $21.79$ & $15.34$ \\
    Full & Pretrained & $4{,}143{,}404{,}816$ & $100.0000$ & $0.3004$ & $40.23$ & $19.61$ \\
    \bottomrule
  \end{tabular}
\end{table*}

%% file: supp/tables/data_scale.tex

\begin{table*}[t]
  \centering
  \caption{Data scale ablation for $\pi_{0.5}$ with full fine-tuning. ``Available'' includes the five held-out appearance variants per task. Every run uses the same validation set ($150$ tasks, $750$ episodes, and $96{,}510$ frames) and $30{,}000$ optimizer updates. The table reports the common endpoint.}
  \label{tab:supp_data_scale}
  \small
  \begin{tabular}{@{}lrrrrrrr@{}}
    \toprule
    \textbf{Scale axis}
      & \textbf{Train tasks}
      & \textbf{Available eps./task}
      & \textbf{Train eps./task}
      & \textbf{Train episodes}
      & \textbf{Train frames}
      & \textbf{Eff. epochs}
      & \textbf{Valid loss} $\downarrow$ \\
    \midrule
    Trajectory density & $150$ & $10$ &  $5$ &   $750$ &  $95{,}445$ & $10.06$ & $0.8755$ \\
    Trajectory density & $150$ & $20$ & $15$ & $2{,}250$ & $274{,}845$ &  $3.49$ & $0.5493$ \\
    Trajectory density & $150$ & $30$ & $25$ & $3{,}750$ & $456{,}510$ &  $2.10$ & $0.4182$ \\
    Trajectory density & $150$ & $40$ & $35$ & $5{,}250$ & $641{,}470$ &  $1.50$ & $0.3361$ \\
    Trajectory density & $150$ & $50$ & $45$ & $6{,}750$ & $823{,}440$ &  $1.17$ & $0.3004$ \\
    \midrule
    Task diversity &  $10$ & $50$ & $45$ &   $450$ &  $53{,}610$ & $17.91$ & $1.0454$ \\
    Task diversity &  $50$ & $50$ & $45$ & $2{,}250$ & $274{,}245$ &  $3.50$ & $0.5574$ \\
    Task diversity & $150$ & $50$ & $45$ & $6{,}750$ & $823{,}440$ &  $1.17$ & $0.3004$ \\
    \bottomrule
  \end{tabular}
\end{table*}